\documentclass[letterpaper]{article}
\usepackage{aaai25}  
\usepackage{aaai25}
\usepackage{times}
\usepackage{helvet}
\usepackage{courier}
\usepackage{graphicx} 
\usepackage{array}
\usepackage{amsmath}
\usepackage{graphicx}
\usepackage{multirow} 
\usepackage{float}
\usepackage{booktabs}
\usepackage{longtable}
\usepackage{booktabs}
\usepackage{multirow}
\usepackage{tabularx}
\usepackage{array}
\usepackage{caption}
\usepackage{color}
\usepackage{pifont}
\usepackage{xcolor} 
\usepackage{colortbl}
\usepackage{url}
\usepackage{aaai25}  
\usepackage{graphicx} 
\usepackage{natbib}  
\usepackage{caption} 
\usepackage{soul}
\usepackage{algorithm}
\usepackage{algorithmic}

\newcommand{\cmark}{\textcolor{green!60!black}{\ding{51}}}%
\newcommand{\xmark}{\textcolor{red!80!black}{\ding{55}}}%

\title{EvoChart: A Benchmark and a Self-Training Approach Towards Real-World Chart Understanding}
\author{
    Muye Huang\textsuperscript{\rm 1,2}, 
    Han Lai\textsuperscript{\rm 1,2}, 
    Xinyu Zhang\textsuperscript{\rm 1,3}, 
    Wenjun Wu\textsuperscript{\rm 1,3}, \\
    Jie Ma\textsuperscript{\rm 2}\thanks{Corresponding author.}, 
    Lingling Zhang\textsuperscript{\rm 1,2}, 
    Jun Liu\textsuperscript{\rm 1,2}
}
\affiliations{
    \textsuperscript{\rm 1}School of Computer Science and Technology, Xi’an Jiaotong University\\
    \textsuperscript{\rm 2}MOE KLINNS Lab, Xi’an Jiaotong University\\
    \textsuperscript{\rm 3}Shaanxi Province Key Laboratory of Big Data Knowledge Engineering\\
    \{huangmuye, hanlai, zhang1393869716, nickjun98\}@stu.xjtu.edu.cn, \\
    \{jiema, zhanglling, liukeen\}@xjtu.edu.cn
}
\begin{document}
\maketitle
\frenchspacing
\begin{abstract}
Chart understanding enables automated data analysis for humans, which requires models to achieve highly accurate visual comprehension. While existing Visual Language Models (VLMs) have shown progress in chart understanding, the lack of high-quality training data and comprehensive evaluation benchmarks hinders VLM chart comprehension. In this paper, we introduce EvoChart, a novel self-training method for generating synthetic chart data to enhance VLMs' capabilities in real-world chart comprehension. We also propose EvoChart-QA, a noval benchmark for measuring models' chart comprehension abilities in real-world scenarios. Specifically, EvoChart is a unique self-training data synthesis approach that simultaneously produces high-quality training corpus and a high-performance chart understanding model. EvoChart-QA consists of 650 distinct real-world charts collected from 140 different websites and 1,250 expert-curated questions that focus on chart understanding. Experimental results on various open-source and proprietary VLMs tested on EvoChart-QA demonstrate that even the best proprietary model, GPT-4o, achieves only 49.8\% accuracy. Moreover, the EvoChart method significantly boosts the performance of open-source VLMs on real-world chart understanding tasks, achieving 54.2\% accuracy on EvoChart-QA.
\end{abstract}
\begin{links}
    \link{Homepage}{https://github.com/MuyeHuang/EvoChart}
\end{links}
\section{Introduction}
Chart Question Answering (CQA) aims to answer specific questions based on the context provided by chart images, enabling automated data analysis, such as the business data reports. This process requires complex chart understanding and visual reasoning skills to interpret various elements, including visual components, text and values. Consequently, CQA tasks have attracted the interest of researchers \cite{DBLP:conf/cvpr/KaflePCK18,DBLP:conf/wacv/MethaniGKK20,DBLP:conf/acl/MasryLTJH22}.

\begin{figure}
    \centering
    \includegraphics[width=1\linewidth]{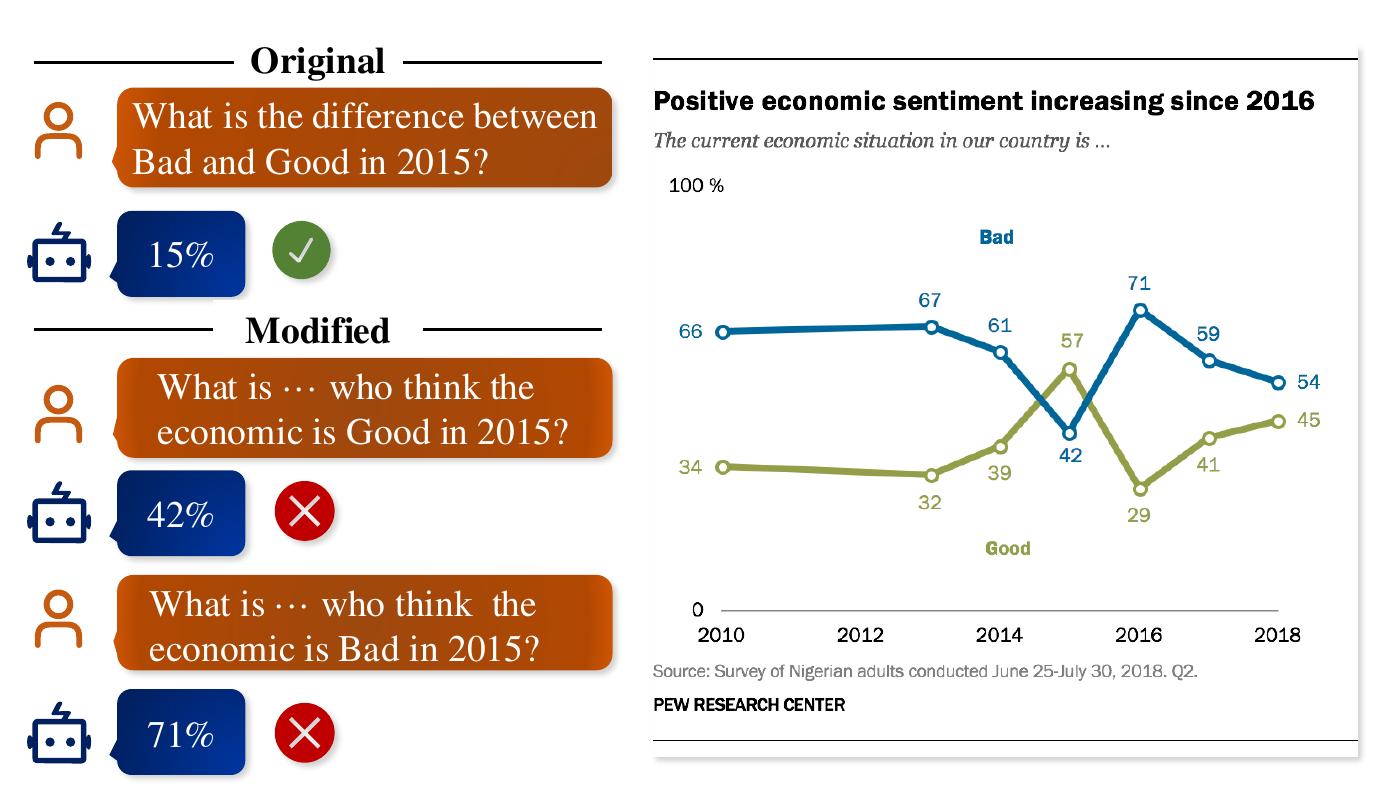}
    \small
    \caption{Case of Modified ChartQA. ``Original" refers to the question from the ChartQA dataset, while ``Modified" refers to our modified version.}
    \label{modified CharQA}
\end{figure}

Recently, VLMs \cite{DBLP:conf/nips/LiuLWL23a,DBLP:conf/nips/Dai0LTZW0FH23,DBLP:journals/corr/abs-2311-07575,DBLP:conf/naacl/0001AJJTT24} have shown significant advancements in general visual capabilities, especially in chart understanding, achieving high scores on the ChartQA dataset \cite{DBLP:journals/corr/abs-2404-16635,DBLP:journals/corr/abs-2312-14238,DBLP:journals/corr/abs-2401-02384}. However, their real-world performance is notably weaker than their ChartQA test set performance. We conducted a test to illustrate this, as shown in Figure \ref{modified CharQA}, we discarded complex reasoning problems in the ChartQA \cite{DBLP:conf/acl/MasryLTJH22} training set and posed 103 basic understanding questions. We then evaluated various VLMs on these questions. The results, presented in the Appendix, show that performance dropped by over 40\% compared to ChartQA scores, even for questions on training set charts. This highlights two points: first, current VLMs are capable of answering some chart-reasoning questions, but they lack a comprehensive understanding of charts. Second, the ChartQA dataset allows models to correctly answer questions without a comprehensive understanding of the charts, leading to an overestimation of the capabilities of current models \cite{DBLP:journals/corr/abs-2406-18521}.

Firstly, the lack of high-quality chart training data is a major reason why current models lack robust chart understanding capabilities. Existing methods \cite{DBLP:conf/acl/MasryLTJH22,DBLP:conf/emnlp/MasryKLHJ23,DBLP:journals/corr/abs-2401-02384} for collecting chart training data fall into two categories: manual annotation and automatic synthesis. Manually annotated data have real-world chart appearances but suffer from coarse granularity and high human costs. Automatically synthesized data offer fine-grained annotations but lack real-world diversity, leading to poor model robustness. Thus, constructing chart datasets is a challenging balance between cost and quality, resulting in a scarcity of high-quality chart training data.

Secondly, the single source of charts and the excessive focus on high-level chart reasoning are primary reasons why the ChartQA dataset provides an overly optimistic estimation of VLM chart understanding capabilities. The ChartQA dataset has only four chart sources, focus on politics and economics. Each source of charts with similar styles, making it prone to overfitting. Additionally, datasets like ChartQA focuses heavily on numerical and logical reasoning, this allows the model to potentially answer questions correctly without a clear understanding of the chart. For example, ``\textit{{What is the difference between Bad and Good in 2015?}}", the model may not explicitly know the values of ``Good" and ``Bad" in 2015, but still has the possibility of answering the question accurately.

To address these challenges, we propose a novel method, EvoChart, for synthesizing high-quality chart datasets with real-world characteristics. We also introduce EvoChart-QA, a carefully crafted benchmark for evaluating chart comprehension in real-world scenarios. 
EvoChart is a multi-stage self-training approach for chart data generation. In each stage, the chart generator produces a batch of synthetic chart data, and the model self-selects and refines the chart data, ensuring that the synthesized data is of high quality for current stage. Subsequently, the model trains on the self-selected data to progress to the next stage. This approach produces both a progressively challenging dataset and a robust chart understanding model.
EvoChart-QA is a benchmark designed for basic chart understanding, featuring 650 charts from 140 real-world websites and 1250 expert-curated questions. The diverse chart styles accurately simulate real-world scenarios, with questions focused on chart understanding. 
Experiments on EvoChart-QA demonstrate that our EvoChart method achieves outstanding performance with 54.2\% accuracy, also exhibits leading performance of 81.5\% on the ChartQA dataset.

Our main contributions are summarized into three folds:
\begin{itemize}
    \item We propose EvoChart, a method that combines chart dataset construction with model self-training, using a multi-stage approach to simultaneously output high-quality chart data and a chart understanding model.
    \item We propose a novel real-world chart basic understanding benchmark, EvoChart-QA, which comprehensively evaluates a model's chart understanding capability through multi-source real-world charts and multi-type manually curated questions.
    \item We conducted extensive experiments on the EvoChart method and EvoChart-QA. Results demonstrate that the EvoChart method significantly outperforms other data synthesis methods, and we also deeply analyze the performance of various VLMs on EvoChart-QA.
\end{itemize}

\section{Related Work}
\subsection{Chart Question Answering Datasets}
Since FigureQA \cite{DBLP:conf/iclr/KahouMAKTB18} pioneered the CQA task, numerous datasets for chart question answering have emerged. Synthetic datasets, such as DVQA \cite{DBLP:conf/cvpr/KaflePCK18}, PlotQA \cite{DBLP:conf/wacv/MethaniGKK20}, RealCQA \cite{DBLP:conf/icdar/AhmedJPSG23}, ChartX \cite{DBLP:journals/corr/abs-2402-12185} and UniChart \cite{DBLP:conf/emnlp/MasryKLHJ23}. This datasets utilize synthetically generated charts or templat-base questions. Generate datasets such as ChartSFT \cite{DBLP:journals/corr/abs-2401-02384}, utilize a mixture of GPT-4 \cite{openai2024gpt4technicalreport}-generated charts and questions. Mixed datasets as ChartQA \cite{DBLP:conf/acl/MasryLTJH22} and Charxiv \cite{DBLP:journals/corr/abs-2406-18521}, the former is a dataset compiled semi-manually with the assistance of templates, while the latter is template-based and requires evaluation by GPT-4o. In contrast, EvoChart-QA focus on real-world scenarios and employ an automated evaluation method that does not necessitate the utilization of GPT-4.

%
\subsection{Visual Language Models on CQA}
VLMs are language models with visual understanding capabilities, and they have numerous applications in CQA tasks. Small VLMs like ChartReader \cite{DBLP:conf/iccv/ChengDH23}, MatCha \cite{DBLP:conf/acl/0001PKPLJACE23}, ScreenAI \cite{DBLP:conf/ijcai/BaechlerSWZMECL24} and UniChart \cite{DBLP:conf/emnlp/MasryKLHJ23} have shown superior performance on tasks like PlotQA and DVQA, highlighting the potential of VLMs in CQA. ChartLlama \cite{DBLP:journals/corr/abs-2311-16483} was a milestone, being the first to apply LLaVa1.5 \cite{liu2024improved} to CQA tasks and achieving impressive performance. Subsequently, works such as ChartPaLI \cite{DBLP:conf/naacl/CarbuneMLABCS24}, ChartInstruct \cite{DBLP:journals/corr/abs-2403-09028}, ChartAst-D \cite{DBLP:journals/corr/abs-2401-02384}, and TinyChart \cite{DBLP:journals/corr/abs-2404-16635} delved into the multimodal alignment and CQA reasoning aspects of VLMs in CQA, achieving remarkable performance. Recently, open-source general VLMs such as Phi3-Vision \cite{abdin2024phi3technicalreporthighly} and Intern-VL2.0 \cite{chen2023internvl}, through large-scale training, have achieved state-of-the-art performance on the ChartQA dataset.

\begin{figure*}[ht]
    \centering
    \includegraphics[width=1\linewidth]{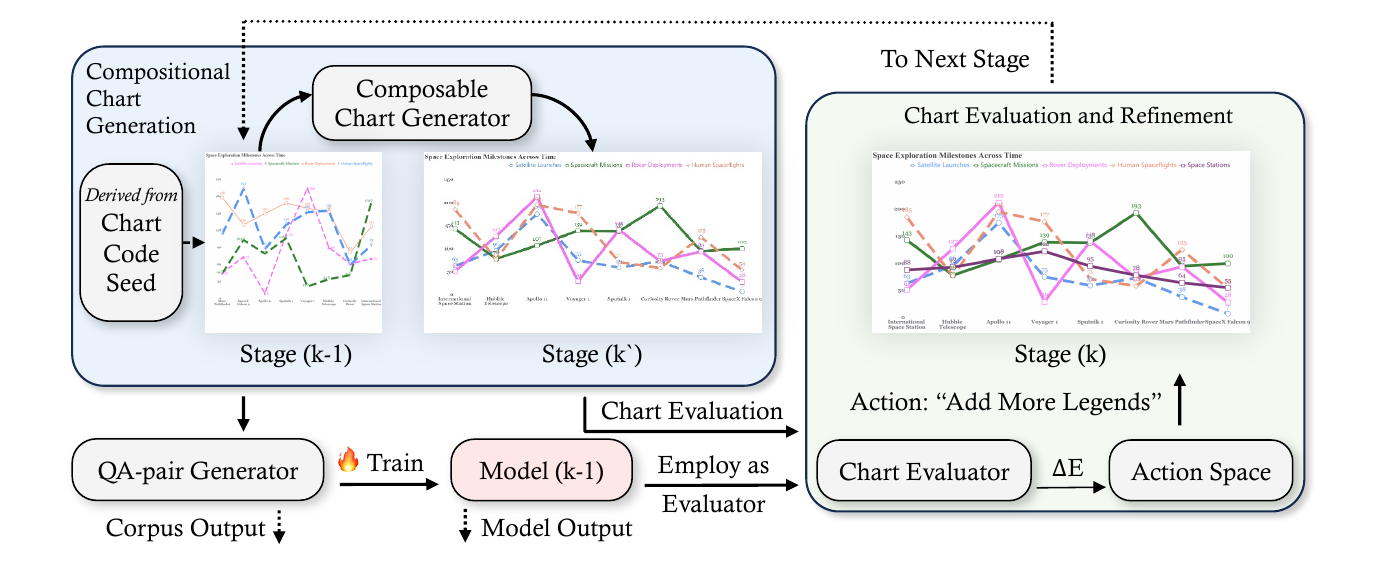}
    \caption{The overview of the proposed EvoChart method. The figure depicts a counterclockwise cyclical self-training process, where the Chart Evaluator of each stage $k$ is trained based on the results of the previous stage $k-1$.}
    \label{overview EvoChart}
\end{figure*}

\subsection{Self-Training Approach}
With the increasing capabilities of language models, numerous researchers have begun to explore the potential of leveraging language models for self-training. GPT3Mix \cite{DBLP:conf/emnlp/YooPKLP21} proved that large language model augmentation of textual corpora is very effective. Further, ReST \cite{DBLP:journals/corr/abs-2308-08998} achieves cost-effective and efficient human preference alignment through a dual-loop self-training approach. Dennis et al. \cite{ulmer2024bootstrappingllmbasedtaskorienteddialogue} obtained new data from the self-talk of multi-role-playing LLM Agents by adding a filtering check mechanism, realizing efficient self-training. Recently, Xu et al. \cite{xu2024interactive} proposed ENVISIONS, which uses a neural-symbolic self-training approach to significantly improve mathematical and logical reasoning abilities without relying on external stronger models or evaluation tools. Inspired by their work, our proposed EvoChart focuses on a scalable self-training process.

\section{EvoChart Method}
We introduce the EvoChart, a unique self-training data synthesis approach that simultaneously produces high-quality training corpus and a high-performance chart understanding model. EvoChart comprises three iterative phases: Compositional Chart Generation for generating charts with diverse appearance, Chart Evaluation and Refinement to select and refine charts suitable for the current stage, and QA-pair generation and training to produce training data and provide a stronger model for the subsequent stage. These phases operate cyclically throughout the construction of EvoChart, as illustrated in Figure \ref{overview EvoChart}. We will explore each of these steps in the following subsections.

\subsection{Compositional Chart Generation}
Compositional chart generation aims to produce high-quality and diverse charts at minimal cost, serving as the core component of EvoChart's construction. Previous approaches \cite{DBLP:journals/corr/abs-2311-16483,DBLP:journals/corr/abs-2401-02384,DBLP:journals/corr/abs-2404-16635} either rely solely on GPT-4 for chart generation result in limited diversity and high costs, or using plotting libraries for random generation often leads to unrealistic themes and styles. To achieve continuous, low-cost, and diverse chart creation, we propose a two-step generation strategy within the compositional chart generation process:

\noindent\textbf{1) Chart Code Seed Generation:} This step serves as the initial phase of chart generation, aiming to produce fundamental chart code containing elements that are difficult to achieve via random processes. These elements include chart themes and appropriate units for the x- and y-axes. Notably, this step is executed only once during the entire chart dataset construction process. As \cite{xu2023symbol} demonstrated the feasibility of large language models for code generation, we employ sophisticated prompt engineering techniques to guide GPT-4 in generating over 25k real-world chart code seeds. Our GPT-4 prompts are based on the following key aspects:

\textbf{Chart Types:} Different chart types are suited for different themes. We focus on the four most prevalent real-world chart types: line charts, bar charts, pie charts, and scatter charts. For each of these chart types, we generate themes that are specifically tailored to their characteristics and use cases.

\textbf{Chart Themes:} Chart themes are challenging to generate through any random process, and manual curation is both costly and prone to domain bias. By utilizing prompting GPT-4, we have generated 25,000 themes across over 200 domains, including politics, economics, technology and everyday life. These themes encompass various titles, units, and other relevant elements.

\textbf{Chart Color Schemes:} Chart color schemes significantly influence the visual appearance of a chart. We employ an automated approach to generate over 200 color palettes, ensuring aesthetically pleasing chart appearances. These color schemes include diverse colors for lines, bars, segments, and backgrounds, among other visual elements.


\noindent\textbf{2) Composable Chart Generator:} The Composable Chart Generator is responsible for producing diverse charts. It is invoked multiple times during the construction process. This step automates the creation of a wide variety of charts by randomly assigning configurations to the chart code. We have defined over a hundred configuration options, each with dozens of potential values. The generator automatically selects these options based on the Code Seed, ensuring diverse chart outputs. Due to the numerous configuration options, we will highlight a few key aspects below:

\textbf{Chart Data:} Numerical data constitutes the core message conveyed by a chart. While randomly generated data may result in excessively volatile values and unrealistic visualizations, we ensure that  the generated chart data is adhered to the specified ranges provided in the Code Seed. This ensures that data values remain within the reasonable bounds defined by the chosen theme.


\textbf{Axis Tick Interval:} Real-world chart creators often omit some labels on axes. For any continuous axis label (e.g., year, month, quarter), we set a 25\% probability of no omission, a 50\% probability of omitting one out of three labels, and a 25\% probability of omitting two out of four labels.

\textbf{Other Configurations:} A multitude of detailed configurations influence chart appearance, including line width, numeric label (position), line style (solid or dashed), bar stacking, axis visibility, font size, font type, and more. We introduce randomness into these configurations through a range of selectable options, and we employ ECharts \cite{DBLP:journals/vi/LiMSSZWZC18} for rendering charts.

\subsection{Chart Evaluation and Refinement}
Chart Evaluation and Refinement enhances the chart images generated by the Compositional Chart Generation process. While chart code seeds can produce diverse charts, refinement remains crucial due to the following reasons: 
1) Random generation can lead to visually similar charts, causing overfitting and reducing the model's generalization ability.
2) Seed-based chart construction may result in poor chart aesthetics, negatively impacting data quality.
To address these issues, we propose two steps: the Chart Evaluator and the Action Space. In the $k$-th stage of data synthesis (Stage-$k$), the Chart Evaluator assigns a multi-dimensional evaluation score $e_k$ to the charts. Based on the difference $\Delta E$ between $e_k$ and the previous score $e_{k-1}$, the Action Space selects an action to modify the charts. The detailed process is described below.

The Chart Evaluator uses the current stage model to assess chart quality, producing a multi-dimensional evaluation score. To avoid hallucinations from questions like ``\textit{{Does this chart have flaws?}}", we assess quality using a directness-based question-answering approach. The Action Space then selects actions to refine the charts based on the evaluation scores. A detailed list of action types is in the Appendix. The evaluation questions and actions are as follows:

\textbf{Is-Chart \& Is-Title-Clear:} These questions check if the chart is correctly rendered. While existing VLMs struggle to comprehend charts in detail, they can still distinguish the names of different chart types. Therefore, we propose the following questions. For example, ``{\textit{Is the image a horizontal bar chart?}}" If the model answers incorrectly, the action is ``Drop." If correct, the action is ``None."




\textbf{Label-Value \& Value-Label:} This question type evaluates chart quality by examining text-value alignment. For example, ``\textit{{What is the value of Medication in May?}}" We generate 10 questions per chart and calculate the average accuracy $e_k$. If $e_k - e_{k-1}$ is significantly positive, the chart may be too simple, prompting a ``value enhancement method." If significantly negative, it may indicate errors or overlaps, prompting the ``Drop" action.

\textbf{Label-Visual \& Visual-Label:} This evaluates visual-text alignment, for example, ``\textit{{What is the bar color of Medication?}}" We generate 10 questions per chart and calculate accuracy $e_k$. If $e_k - e_{k-1}$ is significantly positive, the chart's visual information may be too simple, prompting a ``visual enhancement method." If significantly negative, it may indicate visual errors, prompting the ``Drop" action.

Through Chart Evaluation and Refinement, we ensure that EvoChart generates accurate and challenging data relative to the current stage model in each stage. This ensures data diversity and simultaneously prevents the EvoChart model from overfitting to the EvoChart Corpus.

\subsection{QA-pairs Generation and Training}
QA-pairs Generation and Training aims to generate chart-based question-answer pairs, incorporating these data into the EvoChart corpus and training the EvoChart model for the next stage. We generate question-answer pairs using various question templates. Notably, we focus on basic chart understanding,  the templates specifically focus on the alignment of visual-text-value information in charts (e.g., extracting values through visual information, extracting visual information through text). Additionally, we generate rich question-CoT pairs using composable vCoTs \cite{rose2024visualchainthoughtbridging} and distinguish Direct from vCoT using Instruct. Since vCoTs solely serve to enhance model comprehension, we only mix in 20\% of vCoT data in the training data. We have established 198 question templates with corresponding answers including Direct and over 500 vCoT templates. We generated 1.6M QA-pairs during the training in 3 stages. A detailed information can be found in the Appendix. 
\section{EvoChart-QA Benchmark}

EvoChart-QA is a comprehensive and challenging benchmark for real-world chart understanding. We carefully selected 625 charts with diverse appearances, all sourced from real-world websites. Then we curated 1250 chart-based understanding questions through human experts. This process ensures that EvoChart-QA accurately reflects real-world scenarios. The comparison between EvoChart-QA and other benchmarks is shown in Table \ref{benchmark good bad}. In the following sections, we will elaborate on the chart selection process, question construction methods, and evaluation metrics used. 


\begin{table}[h]
\centering
\small
\begin{tabular}{@{\extracolsep{\fill}} p{1.2cm} | p{0.6cm} p{0.7cm} p{0.9cm} p{0.8cm} p{0.7cm} p{0.7cm} }
\toprule
\multirow{2}{*}{\textbf{Name}} & \textbf{Real Data} & \textbf{Real Chart} & \textbf{Open Vocab} & \textbf{Human Query} & \textbf{Multi Souce} & \textbf{Flex Eval} \\
\midrule
FigureQA & \xmark & \xmark & \xmark & \xmark & \xmark & \xmark \\ 
DVQA & \xmark & \xmark & \xmark & \xmark & \xmark & \xmark \\ 
PlotQA & \cmark & \xmark & \cmark & \xmark & \xmark & \xmark \\ 
ChartQA & \cmark & \cmark & \cmark & \cmark & \xmark & \xmark \\ 
CharXiv & \cmark & \cmark & \cmark & \cmark & \xmark & \xmark \\ 
Ours & \cmark & \cmark & \cmark & \cmark & \cmark & \cmark \\ 
\bottomrule
\end{tabular}

\caption{Comparison with different benchmarks}
\label{benchmark good bad}
\end{table}

\subsection{Chart Selection}
To enable EvoChart-QA to emulate real-world chart understanding scenarios, all charts in our dataset are carefully selected by human experts. Specifically, we crawled 1,000 charts from 140 different websites. Human experts then filtered out images with ambiguous meanings or damages, resulting in a final dataset of 625 valid images. These images include line charts, bar charts, pie charts, and scatter plots. Examples and sources of all images are detailed in the Appendix.

\subsection{Question Construction}
We focus on chart basic understanding questions. Following the definitions of prior researchers \cite{DBLP:conf/wacv/KafleSPCK20,DBLP:conf/wacv/MethaniGKK20}, we concentrate on data and structural retrieval questions. Specifically, we categorize the problems into two types during manual construction: Direct Retrieval and Complex Retrieval. Direct Retrieval questions focus on understanding the image and directly extracting its content, while Complex Retrieval questions emphasize performing multiple visual reasoning steps on the chart.
 
\noindent\textbf{1) Direct Retrieval.}
Direct Retrieval aims to directly extract elements from a chart based on the question. To comprehensively assess the model's ability to extract various elements from charts, we categorize chart elements into three types: label, value, and visual. Label elements refer to textual content in the chart, such as chart title, axis labels, etc. Value elements refer to data values conveyed by the chart, which are displayed or implicitly provided based on the chart author's intention. Visual elements refer to all visual descriptions in the chart, such as line color, largest segment, etc. For example: ``\textit{{What is the value of the green dashed line in 2015?}}" 
Although Direct Retrieval involves only extracting elements from charts, it remains a challenging task. On the one hand, real-world charts often exhibit non-standard variations. For example, they may include rich text such as numerous logos inserted in the image, or they may combine multiple chart forms within a single chart to facilitate expressions. On the other hand, questions posed by real-world users may contain ambiguous expressions. For example, ``\textit{{Which country's total GDP is represented by the bar in the lightest shade of blue?}}" This visual description is ambiguous, but it is a clearly question for human observers.

\begin{table*}[ht]
\centering
\small{
\begin{tabular}{p{2.7cm} *{1}{>{\centering\arraybackslash}p{0.5cm}}*{2}{>{\centering\arraybackslash}p{0.58cm}} *{1}{>{\centering\arraybackslash}p{0.5cm}}*{2}{>{\centering\arraybackslash}p{0.58cm}} *{1}{>{\centering\arraybackslash}p{0.5cm}}*{2}{>{\centering\arraybackslash}p{0.58cm}} *{1}{>{\centering\arraybackslash}p{0.5cm}}*{2}{>{\centering\arraybackslash}p{0.58cm}} *{1}{>{\centering\arraybackslash}p{0.5cm}}*{2}{>{\centering\arraybackslash}p{0.58cm}}}
\toprule
\multirow{2.5}{*}{\textbf{Model}} & \multicolumn{3}{c}{\textbf{Line}} & \multicolumn{3}{c}{\textbf{Bar}} & \multicolumn{3}{c}{\textbf{Pie}} & \multicolumn{3}{c}{\textbf{Scatter}} & \multicolumn{3}{c}{\textbf{Overall}} \\ 
\cmidrule(lr){2-4} \cmidrule(lr){5-7} \cmidrule(lr){8-10} \cmidrule(lr){11-13} \cmidrule(lr){14-16}
                       & \textbf{Dir.} & \textbf{Comp.} & \textbf{All} & \textbf{Dir.} & \textbf{Comp.} & \textbf{All} & \textbf{Dir.} & \textbf{Comp.} & \textbf{All} & \textbf{Dir.} & \textbf{Comp.} & \textbf{All}& \textbf{Dir.} & \textbf{Comp.} & \textbf{All} \\ 
\midrule
\multicolumn{16}{c}{\textit{\textbf{Proprietary Models}}} \\
\midrule
Gemini-1.5-Flash         &26.7    &17.1   & \cellcolor{gray!10}25.0    &28.4   &20.7     & \cellcolor{gray!10}26.8    &41.9    &22.9     & \cellcolor{gray!10}33.8    &33.5    &\underline{19.4}    & \cellcolor{gray!10}29.3    & 30.5   & 20.3    & \cellcolor{gray!10}27.9    \\
Gemini-1.5-Pro          &\underline{42.1}     &21.4     & \cellcolor{gray!10}\underline{38.5}    & 28.4   & 20.7    & \cellcolor{gray!10}26.8    & 41.9   &22.9    &\cellcolor{gray!10} 33.8    & 33.5  & \underline{19.4}    & \cellcolor{gray!10}29.3    & 36.0   & 21.2    &\cellcolor{gray!10} 32.2    \\
Qwen-VL-Plus        &22.7    &11.4     & \cellcolor{gray!10}20.8    &28.8   &13.8     & \cellcolor{gray!10}25.5    & 33.3    &9.4     &\cellcolor{gray!10} 23.1    &27.2   &13.4    & \cellcolor{gray!10}23.1    & 27.0   &  11.9   & \cellcolor{gray!10}23.1     \\
Qwen-VL-Max         & 35.5   &  17.1   & \cellcolor{gray!10}32.2    & 44.4   & 23.0   & \cellcolor{gray!10}39.8    & 48.1   & 25.0    &\cellcolor{gray!10} 38.2    & 33.5  & \underline{19.4}   & \cellcolor{gray!10}29.3    & 39.9   & 21.6    &\cellcolor{gray!10} 35.2    \\
GPT-4-turbo                 & 40.0   & \underline{25.7}    & \cellcolor{gray!10}37.5    & \underline{44.7}   & \underline{29.9}   & \cellcolor{gray!10}\underline{41.5}    &  \textbf{55.0}  &  \underline{34.4}   & \cellcolor{gray!10}\underline{46.2}    &  \underline{46.8}  &  14.9   &\cellcolor{gray!10} \underline{37.3}    &  \underline{44.8}  &  \underline{27.2}   & \cellcolor{gray!10}\underline{40.3}    \\
 GPT-4o                 & \textbf{52.7}   &  \textbf{32.9}   & \cellcolor{gray!10}\textbf{49.2}    & \textbf{52.7}   &  \textbf{44.8}   & \cellcolor{gray!10}\textbf{51.0}    &  \underline{53.5}   & \textbf{49.0}    & \cellcolor{gray!10}\textbf{51.6}    & \textbf{56.3}   & \textbf{23.9}    & \cellcolor{gray!10}\textbf{46.7}    &  \textbf{53.4}  &  \textbf{39.1}   & \cellcolor{gray!10}\textbf{49.8}    \\
\midrule
\multicolumn{16}{c}{\textit{\textbf{Open-source Models}}} \\
\midrule
Phi3-Vision-4B         & \underline{43.3}   & \underline{27.1}    & \cellcolor{gray!10}\underline{40.5}    & \underline{47.9}   & 27.6    & \cellcolor{gray!10}\underline{43.5}    & 33.3   & \underline{27.1}   & \cellcolor{gray!10}30.7    & \underline{50.0}   & 14.9    & \cellcolor{gray!10}\underline{39.6}    & \underline{44.6}   & 24.7    & \cellcolor{gray!10}\underline{39.5}    \\
QwenVL-Chat-7B       & 20.6    &  17.1    & \cellcolor{gray!10}20.0     & 18.2   & 9.2    & \cellcolor{gray!10}16.2    & 31.0   & 14.6    &  \cellcolor{gray!10}24.0   & 24.7   & 11.9    & \cellcolor{gray!10}20.9    & 21.9   &  13.1   &  \cellcolor{gray!10}19.7  \\
LlaVa1.6-Vicuna-7B       & 25.8   &  14.3   & \cellcolor{gray!10}23.8     &  24.9  &  19.5   & \cellcolor{gray!10}23.8    & 38.0   &  15.6   & \cellcolor{gray!10}28.4    & 21.5   & 11.9    & \cellcolor{gray!10}18.7    & 26.5   & 15.6    & \cellcolor{gray!10}23.7    \\
Intern-VL-2.0-8B       & 38.5   & \underline{27.1}    & \cellcolor{gray!10}36.5    & 45.7   & \underline{29.9}   & \cellcolor{gray!10}42.2    & \underline{43.4}   & \underline{27.1}    & \cellcolor{gray!10}\underline{36.4}    &  44.9  & \textbf{22.4}    & \cellcolor{gray!10}38.2    & 42.7    & \underline{26.9}    & \cellcolor{gray!10}38.6    \\
Llama3-Next-8B       & 20.3   & 5.7     & \cellcolor{gray!10} 17.8    & 22.4   & 16.1    & \cellcolor{gray!10}21.0    & 24.0   & 21.9    &   \cellcolor{gray!10}23.1  & 20.9   & 16.4    & \cellcolor{gray!10}19.6   & 21.6   & 15.6    & \cellcolor{gray!10}20.1    \\
CogVLM2-19B       & 24.8   & 11.4    & \cellcolor{gray!10}22.5    & 28.8   & 10.3    & \cellcolor{gray!10}24.8    & 27.9   & 5.2     & \cellcolor{gray!10}18.2    & 24.7   & 7.5      & \cellcolor{gray!10}19.6    & 26.6   & 8.4     & \cellcolor{gray!10}21.9  \\
LlaVa1.6-YI-34B       & 5.8    &  10.0   & \cellcolor{gray!10}6.5      & 7.7    &  4.6    & \cellcolor{gray!10}7.0     & 13.2   & 5.2     & \cellcolor{gray!10}  9.8   & 9.5    & 6.0    &\cellcolor{gray!10} 8.4     & 8.1    & 6.2     &\cellcolor{gray!10}7.6    \\
 Intern-VL-2.0-40B      & \textbf{53.3}   & \textbf{42.9}    & \cellcolor{gray!10}\textbf{51.5}    & \textbf{54.3}   &\textbf{37.9}     & \cellcolor{gray!10}\textbf{50.7}    & \textbf{55.8}    & \textbf{37.5}    & \cellcolor{gray!10}\textbf{48.0}    & \textbf{51.9}    & \underline{20.9}    &\cellcolor{gray!10} \textbf{42.7}    & \textbf{53.8}   & \textbf{35.3}    &\cellcolor{gray!10} \textbf{49.0}    \\
\midrule
\multicolumn{16}{c}{\textit{\textbf{Chart Expert Models}}} \\
\midrule
ChartLlama-13B             &7.3      &4.3      &\cellcolor{gray!10} 6.8     & 7.3    & 10.3    &\cellcolor{gray!10} 8.0    & 21.7    & 6.2     & \cellcolor{gray!10}15.1    & 13.9   &  6.0   & \cellcolor{gray!10}11.6    & 10.4    &  6.9    &\cellcolor{gray!10} 9.5   \\
ChartAst-S-13B             & 12.4   &  12.9   & \cellcolor{gray!10}12.5    & 14.4   & 14.9    & \cellcolor{gray!10}14.5    & 14.7   & 7.3     & \cellcolor{gray!10}11.6    &  13.9  &  12.0   & \cellcolor{gray!10}12.0    & 13.7   &  10.6   & \cellcolor{gray!10}12.9    \\
ChartIns-Llama2-7B   & 17.9   & 11.4    & \cellcolor{gray!10}16.8    & 16.0   & 19.5    & \cellcolor{gray!10}16.8    & 27.1   & 13.5    &\cellcolor{gray!10} 21.3    & 13.9   &  9.0    &\cellcolor{gray!10} 12.4    &  17.8   &  13.8   & \cellcolor{gray!10}16.8    \\
ChartIns-FlanT5-3B   & 23.6   & 24.3    & \cellcolor{gray!10}23.8    & 28.4   &  16.1   & \cellcolor{gray!10}25.8    &  \underline{40.3}  &  19.8   & \cellcolor{gray!10}31.6    & 13.9   &  \underline{19.4}   & \cellcolor{gray!10}15.6    & 25.9   &  19.7    &\cellcolor{gray!10} 24.3    \\
ChartGemma-2B             & \underline{33.9}   & \underline{25.7}    &\cellcolor{gray!10} \underline{32.5}    & \underline{29.1}  & \underline{25.3}   & \cellcolor{gray!10}\underline{28.2}    & 36.4   & \underline{30.2}    & \cellcolor{gray!10}\underline{33.8}    & 32.9  & 16.4    & \cellcolor{gray!10}28.0    & \underline{32.5}   & \underline{25.0}    & \cellcolor{gray!10}\underline{30.6}    \\
TinyChart-3B              & 24.5   &  15.7   &\cellcolor{gray!10} 23.0    & 28.4   & 17.2    & \cellcolor{gray!10}26.0    & 33.3   &  15.6   &\cellcolor{gray!10} 25.8    & \underline{33.5}   & 17.9    &\cellcolor{gray!10} \underline{28.9}    & 28.6   & 16.6   & \cellcolor{gray!10}25.5    \\
 EvoChart-4B        &\textbf{62.1}   &  \textbf{32.9}   & \cellcolor{gray!10}\textbf{57.0}    &  \textbf{62.3}  &  \textbf{33.3}   &\cellcolor{gray!10} \textbf{56.0}    & \textbf{64.3}   & \textbf{30.2}   &\cellcolor{gray!10} \textbf{49.8}    & \textbf{55.1}   & \textbf{37.3}    & \cellcolor{gray!10}\textbf{49.8}    & \textbf{61.3}   &   \textbf{33.1}  & \cellcolor{gray!10}\textbf{54.2}    \\
\bottomrule
\end{tabular}
}

\caption{Experimental results on EvoChart-QA using various open-source or proprietary models. Due to space constraints, abbreviations are used: Dir. refers to Direct, Comp. refers to Complex.}
\label{mainresult}

\end{table*}

\noindent\textbf{2) Complex Retrieval.}
Complex Retrieval involves querying information within a chart using complex, multi-step descriptions. Compared to Direct Retrieval, Complex Retrieval focuses on understanding the relative positions of elements within the chart. For example, ``\textit{{In the chart, the third bar to the left of the longest red bar represents the GDP of which country?}}". Complex retrieval poses novel challenges for chart comprehension. This is because the descriptive information in complex retrieval relies entirely on the chart itself, requiring the model to have a comprehensive and clear understanding of the chart. For example, comprehending the previously mentioned ``\textit{{the longest red bar}}" is entirely based on the extraction of information from the chart's content. Furthermore, this also necessitates the model to possess sophisticated visual reasoning capabilities, such as understanding ``\textit{{the third bar from the left}}" which demands visually-grounded inference.

\subsection{Evaluation Metrics}
We designed a automatic evaluation method for EvoChart-QA, combining flex and strict approaches, to fairly evaluate answer correctness. In EvoChart-QA construction, we label questions as ``Strict" or ``Flex" and use the corresponding Strict or Flex approach to evaluate correctness. For ``Strict" type questions, answers have a definite value, such as numerical or textual values explicitly labeled in the chart. We employ a zero-tolerance approach for judging these questions. For ``Flex" type questions, answers have estimated values, such as unlabeled numerical values. We employ a 5\% tolerance approach to judge these questions. Finally, we employ average accuracy to evaluate the model's performance. In contrast to our metrics, previous methods like ChartQA allowed a 5\% tolerance for any numerical answer, leading to an optimistic estimation of model outputs. For example, years are numerical answers, and 1995 and 2008 would fall within the 5\% tolerance in previous evaluation metrics, and our method does not exhibit this error.

\section{Experiments}
\subsection{Setup}

\noindent\textbf{Datasets.}
To comprehensively evaluate the effectiveness of EvoChart, we chose to test it on both ChartQA and EvoChart-QA. ChartQA \cite{DBLP:conf/acl/MasryLTJH22} is a dataset with two subsets: ``Augment", which is machine-generated, and ``Human", which is manually curated.  ``Augment" focuses on element extraction tasks within machine-synthesized images, while ``Human" emphasizes complex numerical and logical reasoning tasks in real-world charts. EvoChart-QA is a novel real-world benchmark that we proposed.

\noindent\textbf{Models.}
We conducted extensive evaluations on both open-source and proprietary models. For open-source models, we tested Phi3-Vision-4B \cite{abdin2024phi3technicalreporthighly}, QwenVL-Chat-7B \cite{bai2023qwenvlversatilevisionlanguagemodel}, LlaVa1.6-Vicuna-7B \cite{liu2024improved}, Intern-VL-2.0-8B \cite{chen2023internvl}, Llama3-Llava-Next-8B \cite{li2024llava}, CogVLM2-19B \cite{wang2023cogvlm}, LlaVa1.6-YI-34B, Intern-VL-2.0-40B, ChartLlama-13B \cite{DBLP:journals/corr/abs-2311-16483}, ChartAst-S-13B \cite{DBLP:journals/corr/abs-2401-02384}, ChartIns-Llama2-7B \cite{DBLP:journals/corr/abs-2403-09028}, ChartIns-FlanT5-3B, ChartGemma-2B \cite{masry2024chartgemmavisualinstructiontuningchart}, and TinyChart-3B \cite{DBLP:journals/corr/abs-2404-16635}. For proprietary models, we tested Gemini-1.5-Flash \cite{geminiteam2024gemini15unlockingmultimodal}, Gemini-1.5-Pro, Qwen-VL-Plus, Qwen-VL-Max, GPT-4-turbo \cite{openai2024gpt4technicalreport}, and GPT-4o. For all models, we employed a zero-shot approach. The specific configurations of all models are provided in the Appendix. 

\noindent\textbf{Settings.}
In EvoChart method, we utilize Phi3-Vision \cite{abdin2024phi3technicalreporthighly} as the initialization model. We conducted a 3-Stage data synthesis and training process, with each Stage undergoing 1 Epoch of SFT with a learning rate of 2e-5 and using cosine learning rate scheduler. All experiments were completed on 4 NVIDIA A800 80G GPUs.

\begin{figure*}
    \centering
    \includegraphics[width=1.0\linewidth]{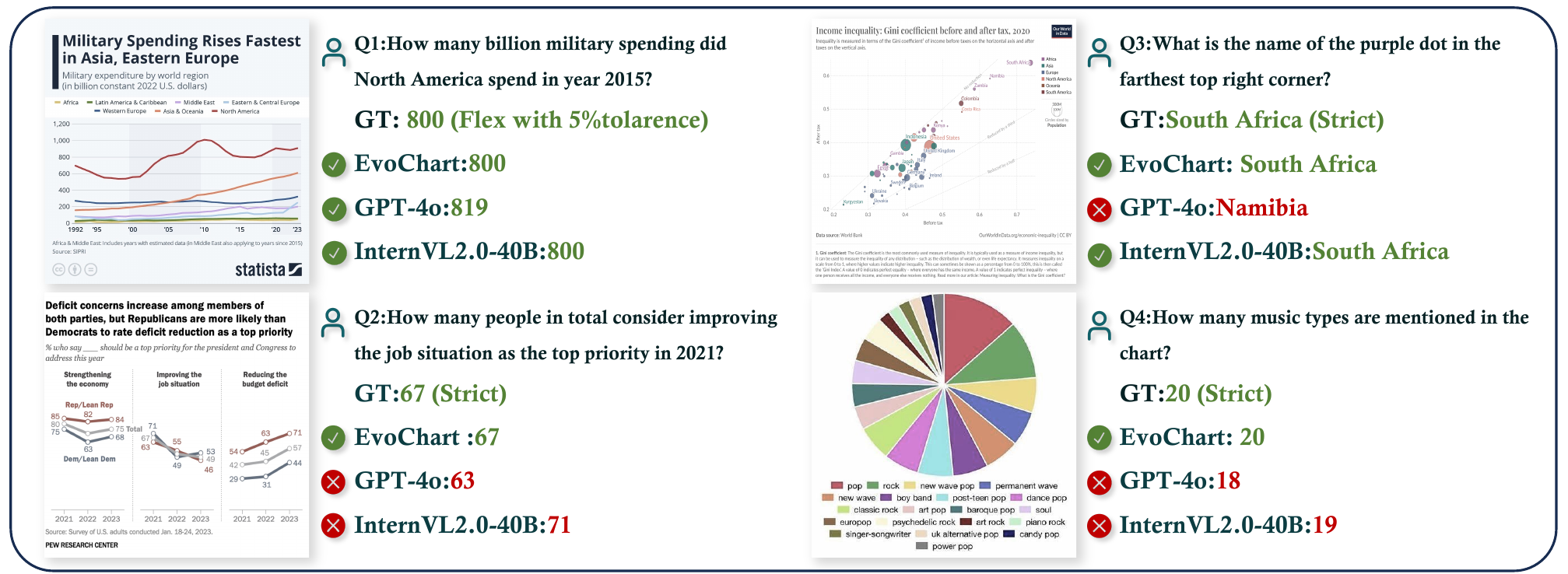}
    \caption{Four cases from the EvoChart-QA Benchmark. Q1 and Q2 are line charts, Q3 is a scatter chart, and Q4 is a pie chart.
}
    \label{case study}
\end{figure*}

\subsection{Experimental Results}
Tables \ref{mainresult} and \ref{ChartQA study} present the performance of EvoChart and other open-source or proprietary VLMs on EvoChart-QA and ChartQA. We elaborate on the experimental results in two aspects: EvoChart-QA and EvoChart.

\noindent \textbf{EvoChart-QA Results.}
All models exhibit relatively poor performance on EvoChart-QA, with accuracies not exceeding 55\%. Among proprietary models, GPT-4o achieves the highest accuracy at 49.8\%. InternVL-2.0-40B demonstrates the strongest performance among open-source general-purpose models, reaching 49.0\%. Within the domain of chart-expert models, our proposed EvoChart method yields the best-performing model, achieving the highest accuracy across all models at 54.2\%. We summarize our findings as follows:

1) EvoChart-QA presents a substantially more challenging benchmark for evaluating basic chart comprehension. Even without involving numerical reasoning or calculation, these models experience a significant performance drop of 30-50\% on EvoChart-QA compared to their scores on ChartQA. This challenge stems from the diversity of charts sourced from 140 websites and the meticulously crafted questions that comprise our dataset.

2) All models demonstrate significantly weaker performance on Complex Retrieval compared to Direct Retrieval, indicating that reasoning over visual information poses a substantially greater challenge than direct extraction of information from charts. Furthermore, nearly all models exhibit lower accuracy on Pie and Scatter chart types compared to their average performance. This suggests that Pie and Scatter charts pose a greater challenge.

3) Open-source general-purpose models exhibit comparable chart comprehension abilities to proprietary models. This suggests that models pretrained on large-scale chart data possess strong generalization capabilities. However, while chart expert models fine-tuned on specific domains achieve impressive scores on the ChartQA dataset, their performance degrades significantly to below 30\% when confronted with the entirely OOD EvoChart-QA dataset.

\begin{table}[t]
\centering
\small
\begin{tabularx}{\linewidth}{X | c c} 
\toprule
\textbf{Model} & \textbf{ChartQA} & \textbf{EvoChart-QA}  \\\midrule
Gemini-1.5-pro &\underline{81.3} &32.2 \\
GPT-4-turbo &62.3 &\underline{40.3}  \\
GPT-4o &\textbf{85.7} &\textbf{49.8}  \\
\midrule
CogVLM2-19B &81.0 &21.9 \\
Phi3-Vision-4B &\underline{81.4} &\underline{38.6}  \\
Intern-VL-2.0-8B &\textbf{81.5} &\textbf{49.0}  \\
\midrule
ChartAst-S-13B &79.9 &12.9  \\
TinyChart-3B &\textbf{83.6} &\underline{25.5}  \\
EvoChart-4B &\underline{81.5} &\textbf{54.2}  \\\bottomrule
\end{tabularx}

\caption{\small Comparison on ChartQA and EvoChart-QA}
\label{ChartQA study}
\end{table}

\noindent\textbf{EvoChart Results.}
Among all proprietary and open-source models evaluated, our proposed EvoChart trained model exhibits significantly superior performance, achieving an accuracy of 54.2\%, surpassing GPT-4o 49.8\%. We have the following observations:

 1) Although the EvoChart model is trained on synthetic data, it achieves SoTA performance on the entirely real-world benchmark EvoChart-QA and exhibits competitive performance on the chart reasoning task ChartQA. This validates the strong generalization ability of the EvoChart. 
 
2) EvoChart primarily focuses on chart basic comprehension. However, as demonstrated in Table \ref{ChartQA study}, EvoChart remains one of the top-performing chart expert models on ChartQA. This is an intriguing finding, suggesting that basic chart comprehension serves as a cornerstone for chart reasoning tasks, and training on basic comprehension can enhance performance in chart reasoning tasks.
 
 3) EvoChart's complex retrieval capabilities are inferior to those of InternVL-2.0-40B and GPT4o. This is reasonable, as these models possess significantly larger scales, which confer an inherent advantage in complex visual extraction and reasoning tasks.
\subsection{Analysis}


\noindent\textbf{EvoChart Ablation Study.}
We conducted comprehensive ablation studies on EvoChart, and the results are summarized in Table \ref{Ablation Study}. We trained and generated EvoChart for 1 to 3 stages. Meanwhile, to verify the effectiveness of Chart Evaluation and Refinement, we set up EvoChart without refinement and trained it for 3 stages, denoted as ``w/o refine stage-3." We observed the following: 

1) As the number of EvoChart method Stages increases, the scale of the EvoChart-Dataset expands, and the performance of the EvoChart steadily improves. This highlights EvoChart’s effectiveness as a self-training approach.

2) Despite having access to a larger training dataset, the ``w/o refine stage-3." exhibits significantly lower performance on EvoChart-QA compared to the complete EvoChart method. This indicates the effectiveness of Chart Evaluation and Refinement in enhancing the model's generalization ability within the EvoChart.

\noindent \textbf{Case Study.}
To further analyze EvoChart and EvoChart-QA, we selected samples from the EvoChart-QA Benchmark for analysis. Figure \ref{case study} presents four cases. More cases are provided in the Appendix. As shown in the figure, overall, EvoChart-QA offers diverse charts and questions, and EvoChart achieves more accurate chart understanding performance compared to GPT4o. Q2 and Q4 demonstrate the effectiveness of our Strict/Flex Metrics. For values explicitly labeled in the image, there should be zero tolerance. However, for estimation questions like Q1, a 5\% tolerance is allowed. Furthermore, Q2 highlights EvoChart's capability for precise chart understanding in complex scenarios. 
\section{Conclusion}
In this paper, we introduce EvoChart and EvoChart-QA: a novel approach for enhancing chart comprehension capabilities through self-training and iterative synthetic data generation, and a meticulously crafted real-world chart comprehension benchmark.  We aim to provide a new avenue for real-world chart understanding through EvoChart and EvoChart-QA. Through extensive experimentation, we expose the limitations of existing VLMs in chart comprehension and validate the effectiveness of our EvoChart method across multiple datasets.  In the future, we will further explore human-free methods in chart comprehension.

\begin{table}[t]
\centering
\small
\begin{tabularx}{\linewidth}{l | c c c c c} 
\toprule
\textbf{Model} & \textbf{Line} & \textbf{Bar}& \textbf{Pie}& \textbf{Scatter}& \textbf{Overall}  \\\midrule
w/  refine stage-1  &53.2 &51.5 &\underline{46.7} &44.9  &50.0 \\
w/  refine stage-2  &\underline{53.5} &\underline{54.2} &\textbf{49.8} &\underline{48.0}  &\underline{52.0} \\
\rowcolor{gray!20} w/  refine stage-3  &\textbf{57.0} &\textbf{56.0} &\textbf{49.8} &\textbf{49.8}  &\textbf{54.2}  \\
w/o refine stage-3  &52.5 &50.7 &43.6 &45.3  &49.0  \\\bottomrule
\end{tabularx}
\caption{\small Ablation Study Results for EvoChart on EvoChart-QA}
\label{Ablation Study}
\end{table}

\section*{Acknowledgments}
This work was supported by National Key Research and Development Program of China (2022YFC3303600), the Key Research and Development Project in Shaanxi Province No. 2022GXLH-01-03, National Natural Science Foundation of China (No. 62137002, 62293553, 62293554, 62450005, 62477036, 62293550, and 62306229), the Shaanxi Provincial Social Science Foundation Project (No. 2024P041), the Natural Science Basic Research Program of Shaanxi (No. 2023-JC-YB-593), the Youth Talent Support Program of Shaanxi Science and Technology Association (20240113), the China Postdoctoral Science Foundation (2024M752585).

\bibliography{aaai25}

\begin{figure*}
    \centering
    \includegraphics[width=1\linewidth]{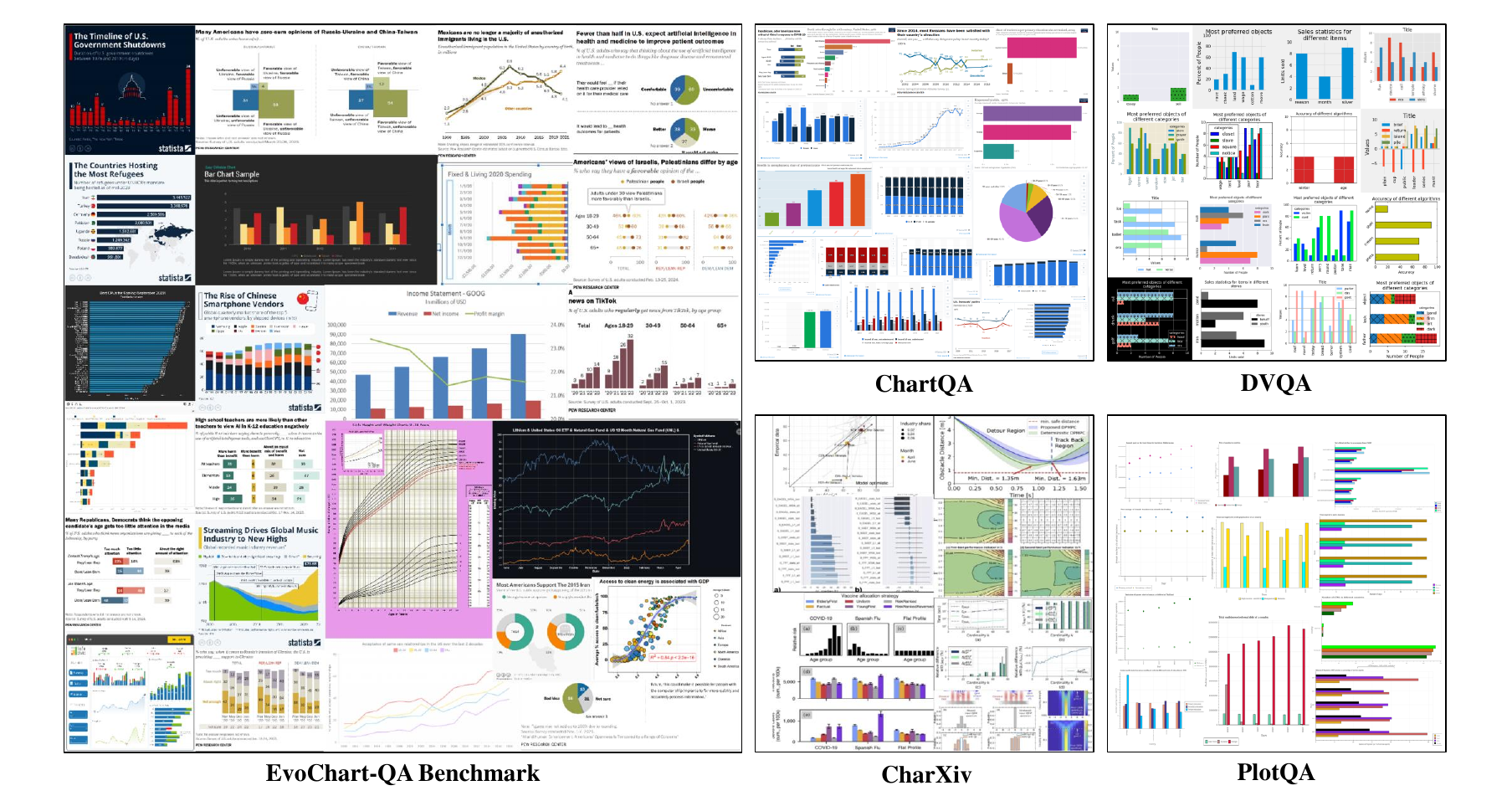}
    \caption{Overview of the EvoChart-QA Benchmark.}
    \label{Overview}
\end{figure*}

\section*{Appendix}

\subsection*{Modified ChartQA}

\subsubsection{1) Result of Modified ChartQA.}
The experimental results on Modified ChartQA using three VLMs, as mentioned in the Introduction, are presented in Table \ref{Modified CharQA}. Even the best-performing Chart Expert model exhibits a performance drop of nearly 30\%.

\begin{table}[h]
\centering
\begin{tabular}{@{}lcc@{}}
\toprule
\textbf{Model} & \textbf{ChartQA-Avg} & \textbf{ChartQA-Modified}  \\ 
\midrule
Gemini-1.5-pro & 81.3 & 18.1 $\downarrow$ \textbf{63.2} \\
GPT-4o & 85.7 & 45.8 $\downarrow$ \textbf{39.9} \\
Phi-3-Vision & 81.4 & 52.6 $\downarrow$ \textbf{28.8} \\
TinyChart & 83.6 & 45.8 $\downarrow$ \textbf{37.8} \\
\bottomrule
\end{tabular}
\caption{Comparison results (\%) on modified CharQA}
\label{Modified CharQA} 
\end{table}

\subsubsection{2) Case of Modified ChartQA.}
\begin{figure*}
    \centering
    \includegraphics[width=1\linewidth]{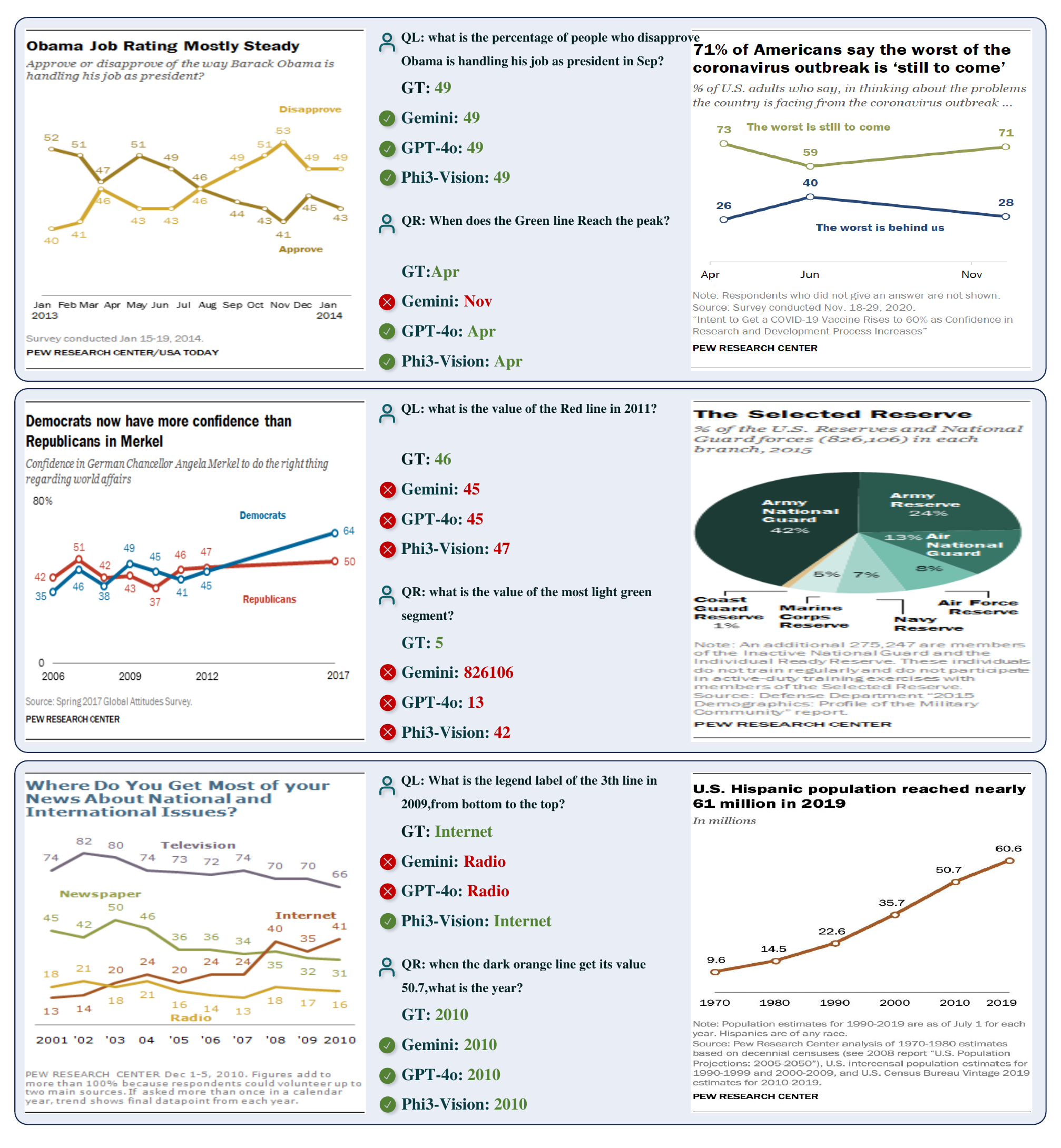}
    \caption{Case 1 of \textbf{Modified ChartQA}, ``QL" indicates that the corresponding image is located on the left side, while ``QR" indicates that the corresponding image is located on the right side.}
    \label{Case of Modified ChartQA1}
\end{figure*}

\begin{figure*}
    \centering
    \includegraphics[width=1\linewidth]{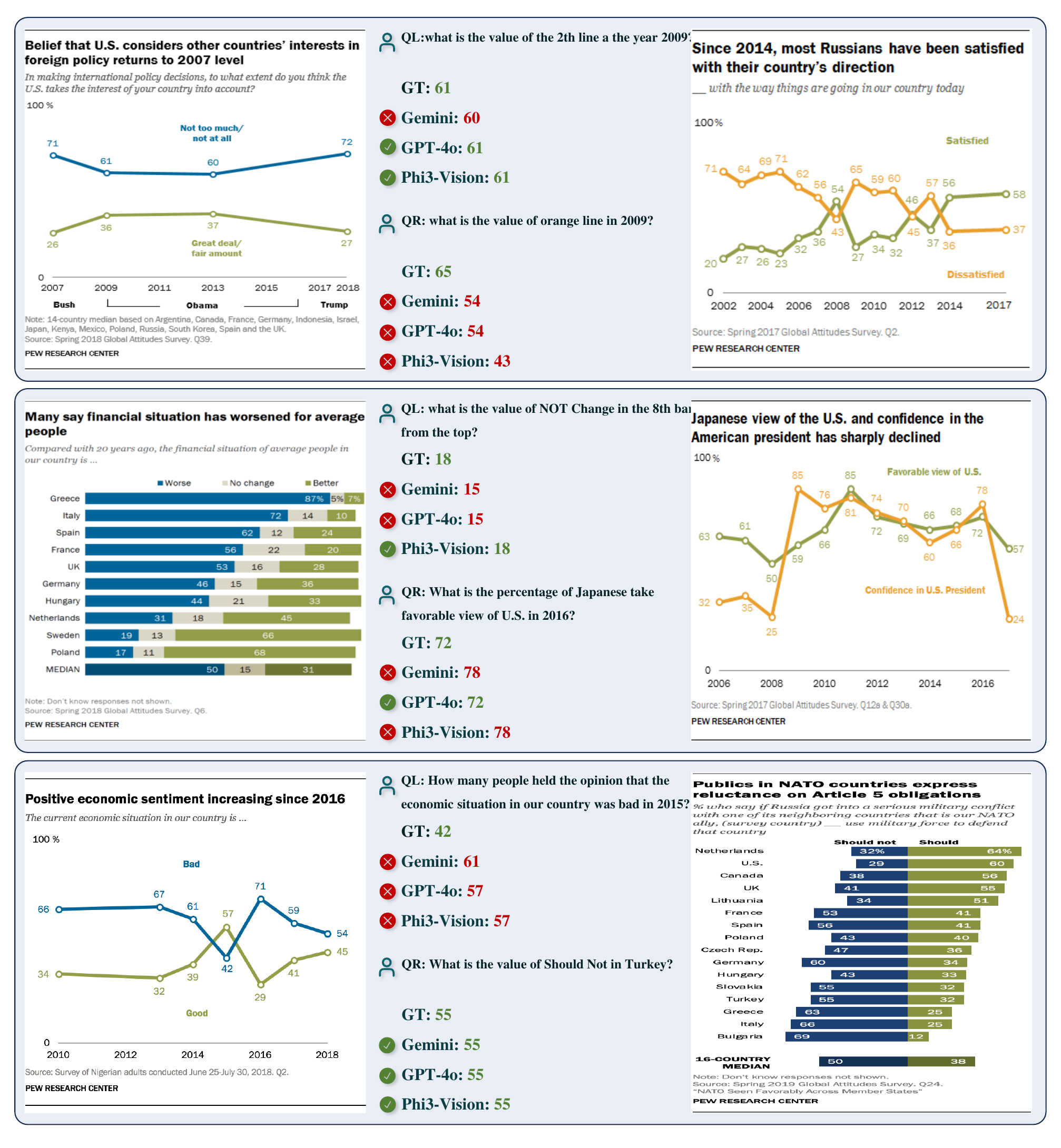}
    \caption{Case 2 of \textbf{Modified ChartQA}, ``QL" indicates that the corresponding image is located on the left side, while ``QR" indicates that the corresponding image is located on the right side.}
    \label{Case of Modified ChartQA2}
\end{figure*}

Examples of the Modified ChartQA cases, as mentioned in the Introduction, are illustrated in Figure \ref{Case of Modified ChartQA1} and Figure \ref{Case of Modified ChartQA2}. The complete set of 103 questions will be released upon publication of this article.

\subsection*{EvoChart-QA Detailed Case}
\begin{figure*}
    \centering
    \includegraphics[width=1\linewidth]{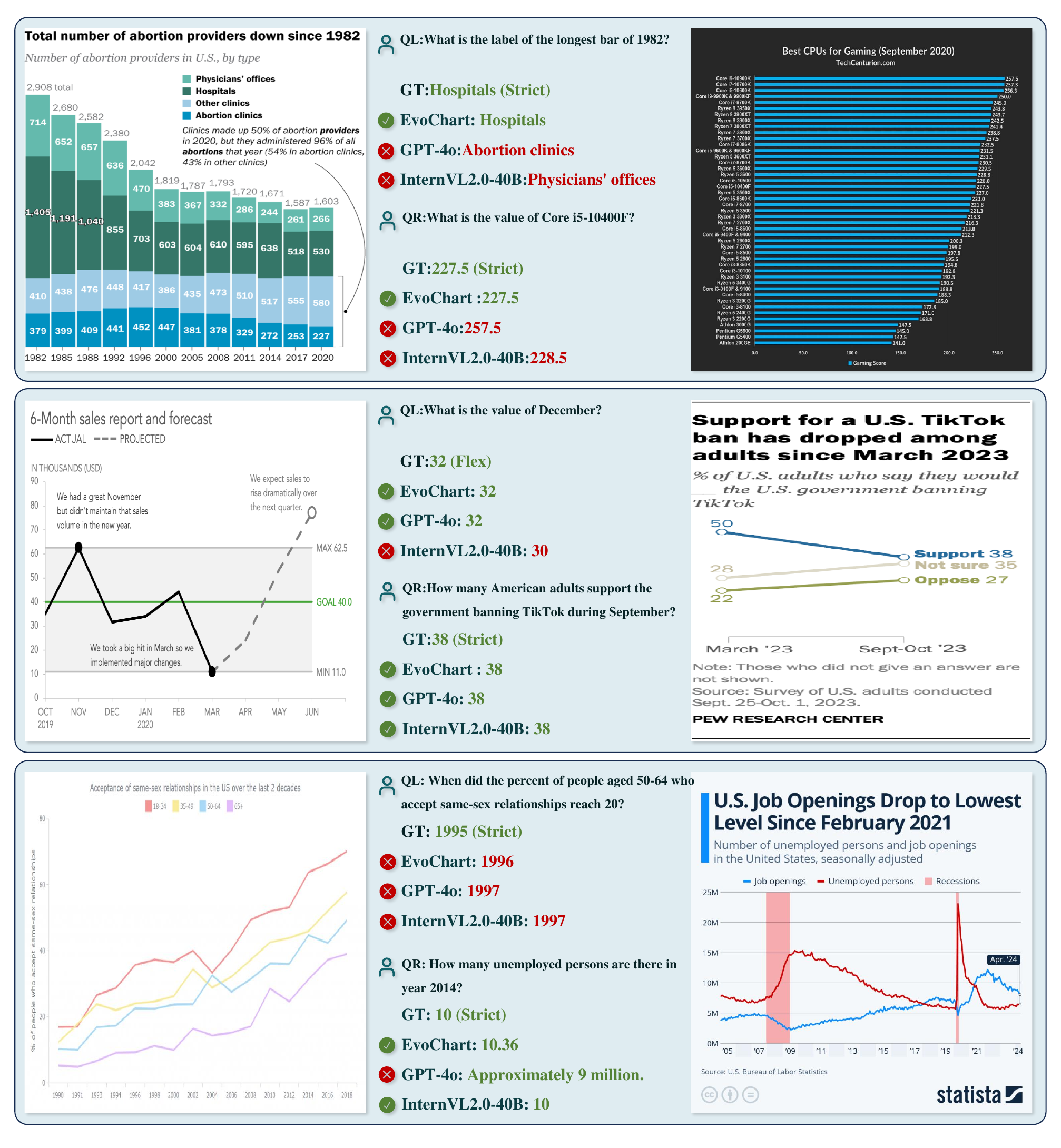}
    \caption{Case 1 of \textbf{EvoChart-QA}, ``QL" indicates that the corresponding image is located on the left side, while ``QR" indicates that the corresponding image is located on the right side.}
    \label{morecase1}
\end{figure*}

\begin{figure*}
    \centering
    \includegraphics[width=1\linewidth]{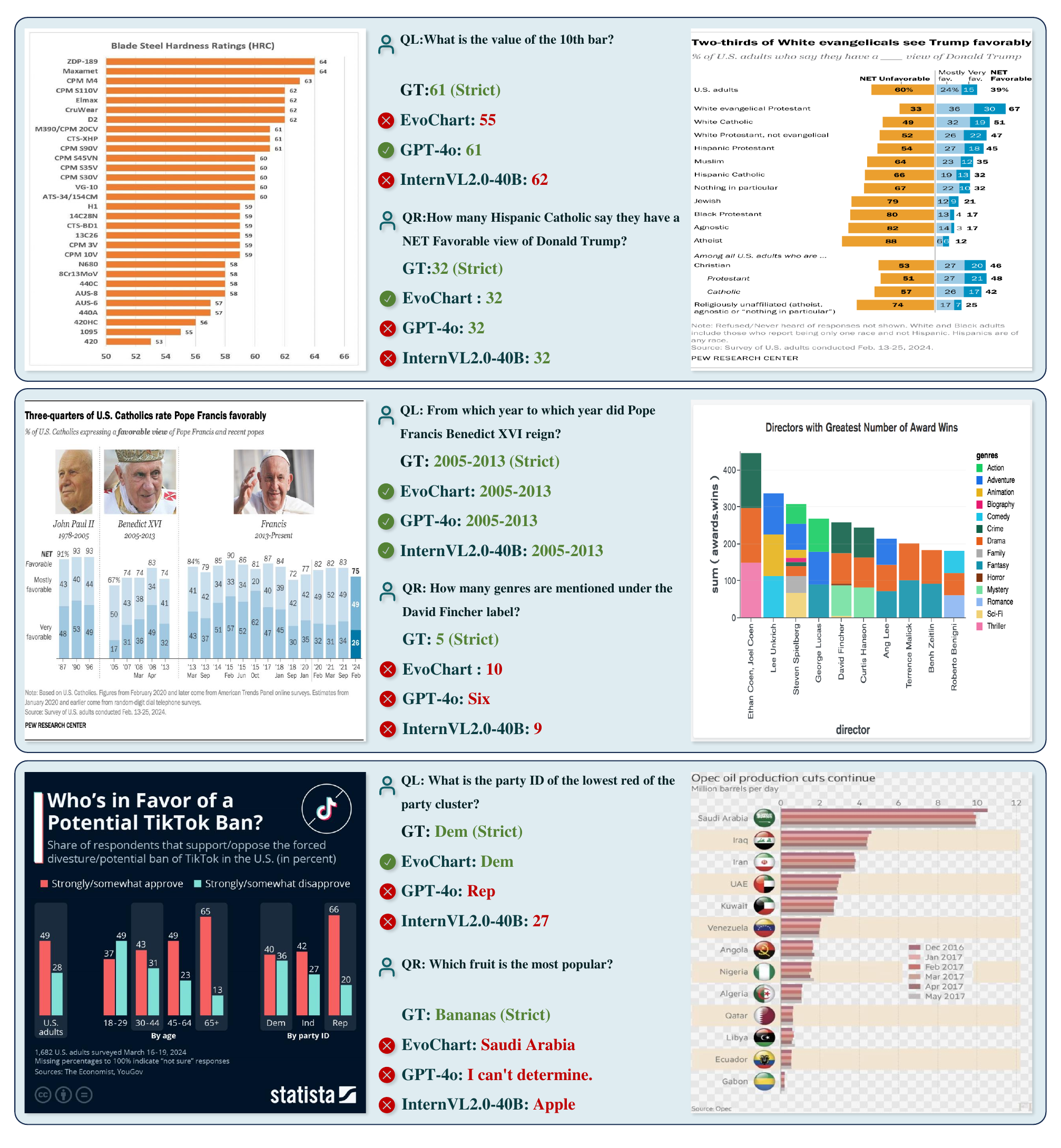}
    \caption{Case 2 of \textbf{EvoChart-QA}, ``QL" indicates that the corresponding image is located on the left side, while ``QR" indicates that the corresponding image is located on the right side.}
    \label{morecase2}
\end{figure*}

\begin{figure*}
    \centering
    \includegraphics[width=1\linewidth]{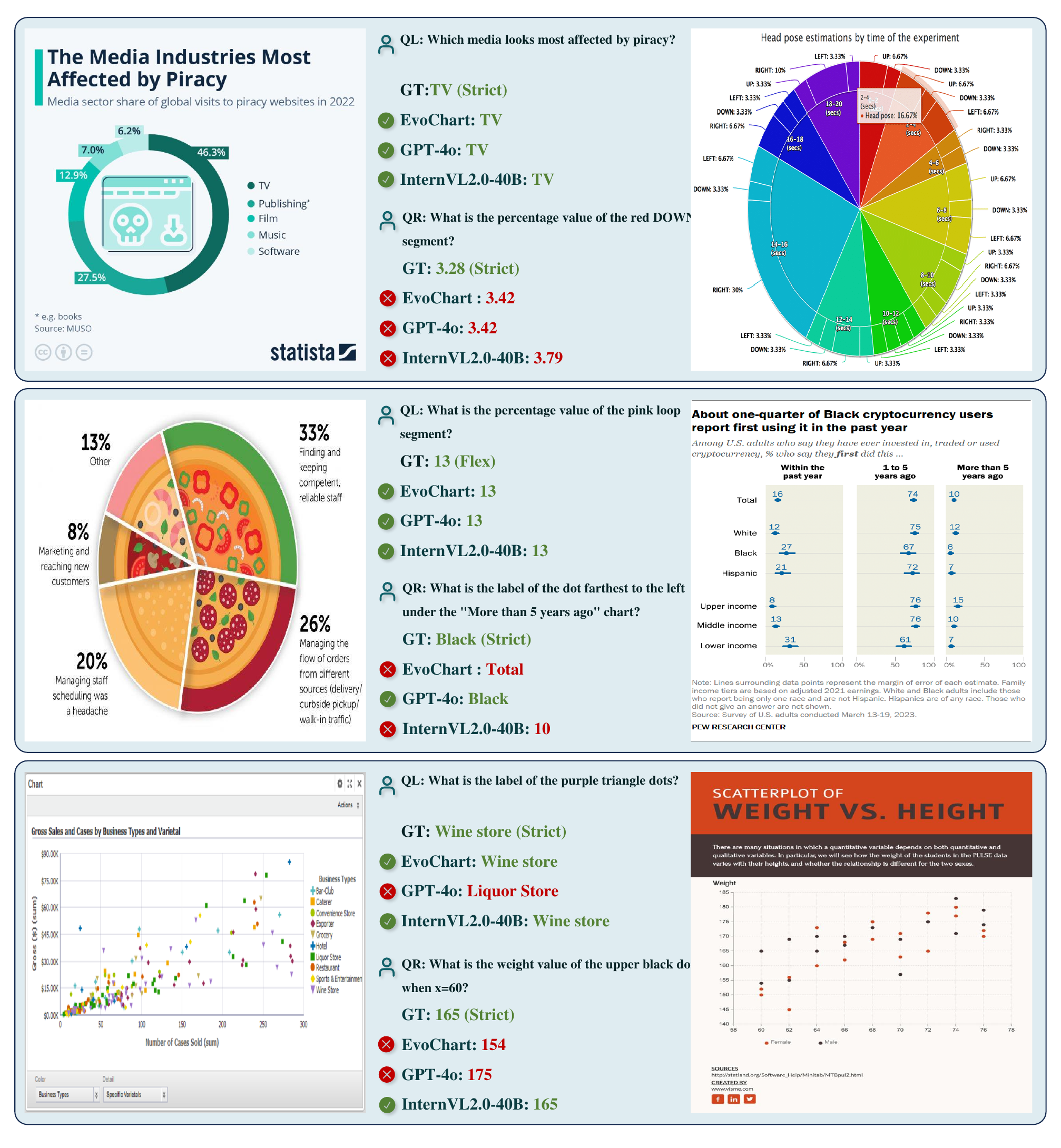}
    \caption{Case 3 of \textbf{EvoChart-QA}, ``QL" indicates that the corresponding image is located on the left side, while ``QR" indicates that the corresponding image is located on the right side.}
    \label{morecase3}
\end{figure*}

An overview of the EvoChart-QA Benchmark is shown in Figure \ref{Overview}. Compared to Charxiv, PlotQA, DVQA, and ChartQA, our proposed EvoChart-QA Benchmark have 140 sources and 1250 manually annotated questions, providing a more realistic evaluation benchmark. This broader scope and meticulous annotation contribute to a more comprehensive and robust assessment of chart comprehension capabilities. Figures \ref{morecase1}, \ref{morecase2}, and \ref{morecase3} present more detailed examples from the EvoChart-QA dataset.

\subsection*{EvoChart Action Space Types}
Table \ref{Action space} presents the categories of actions within EvoChart. The three types correspond to the question types mentioned in the main text. Specifically, VaEM represents Value Enhancement Method, and ViEM represents Visual Enhancement Method. For each question, there is a possibility of dropping its corresponding chart. For questions where VaEM or ViEM is selected, a random action will be chosen to modify the corresponding chart.

\subsection*{Detailed Experiments settings}
For all open-source general-purpose VLMs, proprietary VLMs, and chart expert models, we employ a zero-shot prompting approach. 

For open-source general-purpose and proprietary models, we use the following prompt: ``\textit{You will play as a chart reading expert. You should ONLY give the answer STRING or NUMBER, without any units. You should Not Give Any Explanation.}" This is because general-purpose models have undergone extensive instruction fine-tuning and alignment with human preferences, thus requiring a more detailed prompt to regulate their output.

For chart expert models, we utilize their respective training instructions. For the EvoChart model, the prompt is as follows: ``\textit{You will play as a chart reading expert. You should just give the answer, without any explanation or units.}" This is because chart expert models have been fine-tuned with specific instructions during their training process.

\begin{table}
\centering
\begin{tabular}{@{}lc@{}}
\toprule
\textbf{Type} & \textbf{Actions}   \\ 
\midrule
\multirow{2}{*}{\parbox{3cm}{Is-Chart \& \\ Is-Title-Clear}} & Drop \\ 
                                   & None \\
\midrule
\multirow{5}{*}{\parbox{3cm}{Label-Value \& \\ Value-Label}} & Drop \\
                                   & None \\
                                   & VaEM: Rand Num \\
                                   & VaEM: More Legends  \\
                                   & VaEM: Change Num-Scale  \\
\midrule
\multirow{6}{*}{\parbox{3cm}{Label-Visual \& \\ Visual-Label}} & Drop \\
                                   & None \\
                                   & ViEM: Shuffle Color \\
                                   & ViEM: Change Axis-Scale  \\
                                   & ViEM: Change Color Schemes  \\
                                   & ViEM: Switch Legend Position  \\
\bottomrule
\end{tabular}
\caption{List of Actions in the EvoChart Action Space}
\label{Action space} 
\end{table}

\subsection*{EvoChart Method Question Template}
We have established 284 distinct QA-Pair Templates, as outlined below:
\begin{itemize}
\small
\item Can you tell me the value of \{legend\_label\} in \{xlabel\}?
\item I'd like to know the value of \{legend\_label\} within \{xlabel\}.
\item Could you provide the value of \{legend\_label\} found in \{xlabel\}?
\item What amount does \{legend\_label\} have in \{xlabel\}?
\item Please specify the value of \{legend\_label\} in the context of \{xlabel\}.
\item What is the value of \{legend\_label\} in \{xlabel\} ?
\item Can you identify the legend label with a value of \{value\_label\} at the position marked by \{xlabel\}?
\item What legend label shows a value of \{value\_label\} at the point \{xlabel\}?
\item Could you tell me which legend label corresponds to the value \{value\_label\} at \{xlabel\}?
\item Which label in the legend has the value \{value\_label\} at the \{xlabel\} position?
\item Identify the legend label with a value of \{value\_label\} at \{xlabel\}, please.
\item Which legend label has a value of \{value\_label\} at the position of \{xlabel\} ?
\item List the values at \{xlabel\} from bottom to top.
\item Give me the values at \{xlabel\} arranged in a list from bottom to top.
\item Could you provide the values at \{xlabel\} in a list, starting from the bottom and going to the top?
\item Please provide a list of the values at \{xlabel\} in order from bottom to top.
\item I'd like the values at \{xlabel\} in a list format, ordered from bottom to top.
\item Provide the values at \{xlabel\} in a list format from bottom to top.
\item What are the data values in ascending order on the x-axis tick right before \{xlabel\}?
\item Can you list the data values from smallest to largest on the x-axis tick just to the left of \{xlabel\}?
\item Please provide the data values sorted from smallest to largest for the x-axis tick immediately preceding \{xlabel\}.
\item Could you tell me the data values from smallest to largest at the x-axis tick just before \{xlabel\}?
\item What are the data values, ordered from smallest to largest, on the x-axis tick directly left of \{xlabel\}?
\item On the x-axis tick immediately to the left of \{xlabel\}, what are the data values from smallest to largest?
\item What legend label features a \{line\_color\} \{line\_style\} line?
\item Which legend key shows a \{line\_color\} \{line\_style\} line?
\item Can you identify the legend label with a \{line\_color\} \{line\_style\} line?
\item Which label in the legend corresponds to a \{line\_color\} \{line\_style\} line?
\item Could you tell me which legend label has a \{line\_color\} \{line\_style\} line?
\item Which legend label has a \{line\_color\} \{line\_style\} line?
\item Can you tell me the line style for \{legend\_label\}?
\item What kind of line style does \{legend\_label\} use?
\item Please specify the line style associated with \{legend\_label\}.
\item How is the line style defined for \{legend\_label\}?
\item What's the type of line style for \{legend\_label\}?
\item What is the line style of \{legend\_label\}?
\item Can you tell me the value of the \{n\}th data point from the left on the \{line\_color\} line that is \{line\_style\}?
\item What is the value of the \{n\}th point from the left on the \{line\_style\} line that is colored \{line\_color\}?
\item Please provide the value of the \{n\}th data point from the left on the \{line\_color\} \{line\_style\} line.
\item Could you specify the value of the \{n\}th point from the left on the \{line\_style\} line in \{line\_color\}?
\item What value does the \{n\}th point from the left have on the \{line\_color\} line with \{line\_style\}?
\item What is the value of the \{n\}th data point from left to right on the \{line\_style\} line of \{line\_color\} color?
\item When the line labeled \{legend\_label\} hits the \{value\_label\} mark at \{xlabel\}, how many lines is it positioned above?
\item At \{xlabel\}, when the \{legend\_label\} line reaches the \{value\_label\} level, how many other lines is it above?
\item How many lines are beneath the \{legend\_label\} line when it reaches \{value\_label\} at \{xlabel\}?
\item When \{legend\_label\} hits the value \{value\_label\} at \{xlabel\}, how many lines are below it?
\item How many lines does the line labeled \{legend\_label\} surpass at \{xlabel\} when it attains the \{value\_label\} value?
\item When the line represented by \{legend\_label\} reaches the value \{value\_label\} at \{xlabel\}, how many lines is this line above?
\item At which x-label does the line represented by \{legend\_label\} reach its highest point?
\item Where does the line indicated by \{legend\_label\} peak on the x-axis?
\item Can you identify the x-label where the line for \{legend\_label\} reaches its maximum value?
\item Which x-label corresponds to the highest point of the line denoted by \{legend\_label\}?
\item At what x-label is the peak of the line marked by \{legend\_label\}?
\item The highest point of the line represented by \{legend\_label\} is at which x-label?
\item Where on the x-axis does the line labeled \{legend\_label\} dip to its lowest value?
\item Can you tell me the x-label where the \{legend\_label\} line hits its minimum?
\item At which point on the x-axis does the \{legend\_label\} line bottom out?
\item What's the x-label where the line for \{legend\_label\} reaches its lowest point?
\item Where along the x-axis does the \{legend\_label\} line find its lowest value?
\item At which x-label does the line represented by \{legend\_label\} reach its lowest point?
\item At which x-label does the line with color \{line\_color\} and style \{line\_style\} reach its lowest point?
\item Where along the x-axis does the \{line\_color\} line, with its \{line\_style\} style, hit the bottom?
\item Can you tell me the x-label where the \{line\_color\} line, designed in \{line\_style\}, drops to its lowest?
\item Which x-label marks the lowest point of the \{line\_color\} line that’s styled as \{line\_style\}?
\item What's the x-label where the \{line\_color\} line, styled in \{line\_style\}, touches its lowest point?
\item Which xlabel does the \{line\_color\} line, styled as \{line\_style\}, reach its bottom?
\item At which xlabel does the \{line\_color\} colored line with \{line\_style\} reach its peak?
\item Can you tell me the value of \{legend\_label\} in \{xlabel\}?
\item I'd like to know the value of \{legend\_label\} within \{xlabel\}.
\item Could you provide the value of \{legend\_label\} found in \{xlabel\}?
\item What amount does \{legend\_label\} have in \{xlabel\}?
\item Please specify the value of \{legend\_label\} in the context of \{xlabel\}.
\item What is the value of \{legend\_label\} in \{xlabel\} ?
\item What is the legend label for the value \{value\_label\} on the \{xlabel\} axis?
\item Identify the legend label that corresponds to the value \{value\_label\} in \{xlabel\}.
\item Which legend label matches the value \{value\_label\} in \{xlabel\}?
\item Find the legend label associated with the value \{value\_label\} in \{xlabel\}.
\item Can you tell me the legend label that has the value \{value\_label\} in \{xlabel\}?
\item Which legend label have value of \{value\_label\} in \{xlabel\} ?
\item Which legend label has a value of \{value\_label\} at the position of \{xlabel\}?
\item At the position of \{xlabel\}, which legend label corresponds to the value \{value\_label\}?
\item Identify the legend label that has a value of \{value\_label\} at the \{xlabel\} position.
\item What legend label holds the value \{value\_label\} at the position indicated by \{xlabel\}?
\item Determine the legend label with a value of \{value\_label\} at the \{xlabel\} location.
\item Which legend label shows a value of \{value\_label\} at the position marked by \{xlabel\}?
\item Provide the values at \{xlabel\} in a list format rating from small to large.
\item Can you list the values at \{xlabel\} from small to large?
\item Please arrange the values at \{xlabel\} in a list from small to large.
\item List the values at \{xlabel\} in ascending order.
\item I'd like to see the values at \{xlabel\} rated from small to large in a list.
\item Can you provide a list of the values at \{xlabel\} from the smallest to the largest?
\item Could you please give the values at \{xlabel\} in a list format, ordered from small to large?
\item What is the value of the \{n\}th data point from bottom to top on the \{border\_type\} border bar of \{line\_color\} color?
\item Could you tell me the value of the \{n\}th data point from the bottom to top on the \{border\_type\} border bar in \{line\_color\}?
\item Please provide the value of the \{n\}th data point from bottom to top on the \{line\_color\} \{border\_type\} border bar.
\item What's the value of the \{n\}th data point from bottom to top on the \{line\_color\} \{border\_type\} border bar?
\item I need the value of the \{n\}th data point from the bottom to the top on the \{border\_type\} border bar of \{line\_color\}.
\item Can you find the value of the \{n\}th data point from bottom to top on the \{line\_color\} border bar of \{border\_type\} type?
\item Please tell me the value of the \{n\}th data point from bottom to top on the \{border\_type\} border bar that is \{line\_color\}.
\item What is the value of the longest \{line\_color\} bar?
\item Can you tell me the value of the tallest \{line\_color\} bar?
\item I need to know the value of the highest \{line\_color\} bar.
\item What is the value of the \{line\_color\} bar with the maximum height?
\item Please provide the value of the largest \{line\_color\} bar.
\item Could you find out the value of the highest \{line\_color\} bar?
\item What is the value of the shortest \{line\_color\} bar?
\item Can you tell me the value of the smallest \{line\_color\} bar?
\item I need to know the value of the least tall \{line\_color\} bar.
\item What is the value of the \{line\_color\} bar with the minimum height?
\item Please provide the value of the \{line\_color\} bar that is the shortest.
\item Could you find out the value of the shortest \{line\_color\} bar for me?
\item Can you tell me the value of \{legend\_label\} in \{xlabel\}?
\item I'd like to know the value of \{legend\_label\} within \{xlabel\}.
\item Could you provide the value of \{legend\_label\} found in \{xlabel\}?
\item What amount does \{legend\_label\} have in \{xlabel\}?
\item Please specify the value of \{legend\_label\} in the context of \{xlabel\}.
\item What is the value of \{legend\_label\} in \{xlabel\} ?
\item What is the legend label for the value \{value\_label\} on the \{xlabel\} axis?
\item Identify the legend label that corresponds to the value \{value\_label\} in \{xlabel\}.
\item Which legend label matches the value \{value\_label\} in \{xlabel\}?
\item Find the legend label associated with the value \{value\_label\} in \{xlabel\}.
\item Can you tell me the legend label that has the value \{value\_label\} in \{xlabel\}?
\item Which legend label have value of \{value\_label\} in \{xlabel\} ?
\item Which legend label has a value of \{value\_label\} at the position of \{xlabel\}?
\item At the position of \{xlabel\}, which legend label corresponds to the value \{value\_label\}?
\item Identify the legend label that has a value of \{value\_label\} at the \{xlabel\} position.
\item What legend label holds the value \{value\_label\} at the position indicated by \{xlabel\}?
\item Determine the legend label with a value of \{value\_label\} at the \{xlabel\} location.
\item Which legend label shows a value of \{value\_label\} at the position marked by \{xlabel\}?
\item Provide the values at \{xlabel\} in a list format rating from small to large.
\item Can you list the values at \{xlabel\} from small to large?
\item Please arrange the values at \{xlabel\} in a list from small to large.
\item List the values at \{xlabel\} in ascending order.
\item I'd like to see the values at \{xlabel\} rated from small to large in a list.
\item Can you provide a list of the values at \{xlabel\} from the smallest to the largest?
\item Could you please give the values at \{xlabel\} in a list format, ordered from small to large?
\item What is the value of the \{n\}th data point from left on the \{border\_type\} border bar of \{line\_color\} color?
\item Could you tell me the value of the \{n\}th data point from left on the \{border\_type\} border bar in \{line\_color\}?
\item Please provide the value of the \{n\}th data point from left on the \{line\_color\} \{border\_type\} border bar.
\item What's the value of the \{n\}th data point from left on the \{line\_color\} \{border\_type\} border bar?
\item I need the value of the \{n\}th data point from left on the \{border\_type\} border bar of \{line\_color\}.
\item Can you find the value of the \{n\}th data point from left on the \{line\_color\} border bar of \{border\_type\} type?
\item Please tell me the value of the \{n\}th data point from left on the \{border\_type\} border bar that is \{line\_color\}.
\item What is the value of the longest \{line\_color\} bar?
\item Can you tell me the value of the tallest \{line\_color\} bar?
\item I need to know the value of the highest \{line\_color\} bar.
\item What is the value of the \{line\_color\} bar with the maximum height?
\item Please provide the value of the largest \{line\_color\} bar.
\item Could you find out the value of the highest \{line\_color\} bar?
\item What is the value of the shortest \{line\_color\} bar?
\item Can you tell me the value of the smallest \{line\_color\} bar?
\item I need to know the value of the least tall \{line\_color\} bar.
\item What is the value of the \{line\_color\} bar with the minimum height?
\item Please provide the value of the \{line\_color\} bar that is the shortest.
\item Could you find out the value of the shortest \{line\_color\} bar for me?
\item How many sectors are there in total in this pie chart?
\item How many segments are there in total in this pie chart?
\item What is the total number of sectors in this pie chart?
\item How many segments in total are present in this pie chart?
\item What is the total count of sectors in this pie chart?
\item How many segments does this pie chart have in total?
\item What is the percentage of '{sector\_label}' in '{series\_label}' on this chart?
\item How much percent does '{sector\_label}' make up in '{series\_label}' on this chart?
\item Can you tell me the percentage of '{sector\_label}' within '{series\_label}' in this chart?
\item What proportion of '{series\_label}' does '{sector\_label}' represent in this chart?
\item How large is the percentage of '{sector\_label}' in the '{series\_label}' shown on this chart?
\item In this chart, what percentage does '{sector\_label}' constitute in '{series\_label}'?
\item What is the number of '{sector\_label}' in '{series\_label}' on this chart?
\item How many '{sector\_label}' are there in '{series\_label}' on this chart?
\item Can you find the number of '{sector\_label}' within '{series\_label}' in this chart?
\item What count of '{sector\_label}' does '{series\_label}' have in this chart?
\item How many instances of '{sector\_label}' are in '{series\_label}' on this chart?
\item In this chart, what is the count of '{sector\_label}' in '{series\_label}'?
\item What percentage of '{series\_label}' is made up by '{sector\_label}' in this chart?
\item What is the proportion of '{sector\_label}' in '{series\_label}' on this chart?
\item Could you specify the fraction of '{sector\_label}' within '{series\_label}' depicted in this chart?
\item What ratio does '{sector\_label}' contribute to '{series\_label}' as shown in this chart?
\item How large is the share of '{sector\_label}' in '{series\_label}' on this chart?
\item In this chart, what part of {series\_label} does '{sector\_label}' represent?
\item What is the percentage of the sector with the \{name\_color\} color?
\item How much percent does the sector with the \{name\_color\} color represent?
\item Can you tell me the proportion of the sector with the \{name\_color\} color?
\item What fraction does the sector with the \{name\_color\} color make up?
\item How large is the percentage of the sector with the \{name\_color\} color?
\item In this chart, what percentage does the sector with the \{name\_color\} color constitute?
\item What is the number of the sector with the \{name\_color\} color?
\item How many sectors are there with the \{name\_color\} color?
\item Can you tell me the count of the sector with the \{name\_color\} color?
\item What is the quantity of the sector with the \{name\_color\} color?
\item How many sectors are labeled with the \{name\_color\} color?
\item In this chart, what is the number of sectors with the \{name\_color\} color?
\item What is the percentage of the sector with the \{name\_color\} color?
\item How much of the sector is represented by the \{name\_color\} color in percentage?
\item Can you tell me the percentage of the sector that is \{name\_color\}?
\item What fraction of the sector is the \{name\_color\} color?
\item How large is the share of the sector with the \{name\_color\} color?
\item What portion of the sector does the \{name\_color\} color represent in percentage?
\item What is the percentage of the largest sector in '{series\_label}' in the pie chart?
\item In the pie chart, what percentage does the largest sector in '{series\_label}' represent?
\item What proportion does the largest sector in '{series\_label}' hold in the pie chart?
\item How much percentage does the largest sector in '{series\_label}' account for in the pie chart?
\item Can you tell me the percentage of the largest sector in '{series\_label}' on the pie chart?
\item What is the share of the largest sector in '{series\_label}' in the pie chart?
\item What percentage of the pie chart does the smallest sector in '{series\_label}' occupy?
\item In '{series\_label}', what is the percentage of the smallest sector in the pie chart?
\item What is the fractional representation of the smallest sector in '{series\_label}' within the pie chart?
\item How much does the smallest sector in '{series\_label}' contribute to the pie chart as a percentage?
\item What is the smallest sector's percentage in the pie chart under '{series\_label}'?
\item Within '{series\_label}', what is the percentage value of the smallest sector in the pie chart?
\item Can you tell me the \{y\_axis\_topic\} of \{x\_value\} \{x\_axis\_topic\} in \{legend\_name\}?
\item What is the \{y\_axis\_topic\} of \{x\_value\} \{x\_axis\_topic\} in \{legend\_name\}?
\item Please provide the \{y\_axis\_topic\} for \{x\_value\} \{x\_axis\_topic\} in \{legend\_name\}.
\item Could you tell me the \{y\_axis\_topic\} for \{x\_value\} \{x\_axis\_topic\} in \{legend\_name\}?
\item What \{y\_axis\_topic\} corresponds to \{x\_value\} \{x\_axis\_topic\} in \{legend\_name\}?
\item Can you provide the \{y\_axis\_topic\} of \{x\_value\} \{x\_axis\_topic\} in \{legend\_name\}?
\item Could you provide the \{x\_axis\_topic\} for a \{y\_value\} \{y\_axis\_topic\} in \{legend\_name\}?
\item What is the \{x\_axis\_topic\} for a \{y\_value\} \{y\_axis\_topic\} in \{legend\_name\}?
\item Can you give the \{x\_axis\_topic\} for a \{y\_value\} \{y\_axis\_topic\} in \{legend\_name\}?
\item Please provide the \{x\_axis\_topic\} corresponding to a \{y\_value\} \{y\_axis\_topic\} in \{legend\_name\}.
\item Could you tell me the \{x\_axis\_topic\} for a \{y\_value\} \{y\_axis\_topic\} in \{legend\_name\}?
\item What \{x\_axis\_topic\} corresponds to a \{y\_value\} \{y\_axis\_topic\} in \{legend\_name\}?
\item How many legend labels are there in the chart?
\item What is the number of legend labels in the chart?
\item In the chart, how many legend labels are present?
\item How many labels are there in the chart's legend?
\item What count of legend labels is shown in the chart?
\item Can you tell how many legend labels are included in the chart?
\item How many different colors of data points are there in the chart?
\item What is the number of different colors of data points in the chart?
\item In the chart, how many different colors of data points can be observed?
\item How many unique colors of data points are present in the chart?
\item What count of different colored data points is shown in the chart?
\item Can you tell how many different colors of data points are in the chart?
\item What is the \{x\_axis\_topic\} value of the \{legend\_name\} at the peak \{y\_axis\_topic\} in this chart?
\item What is the \{x\_axis\_topic\} value of the \{legend\_name\} at the peak \{y\_axis\_topic\} in this chart?
\item What \{x\_axis\_topic\} value corresponds to the peak \{y\_axis\_topic\} for the \{legend\_name\} in this chart?
\item In this chart, what is the \{x\_axis\_topic\} value when \{legend\_name\} reaches the peak \{y\_axis\_topic\}?
\item At the peak \{y\_axis\_topic\} for \{legend\_name\} in this chart, what is the \{x\_axis\_topic\} value?
\item What is the \{x\_axis\_topic\} value when \{legend\_name\} has the peak \{y\_axis\_topic\} in this chart?
\item In this chart, what \{x\_axis\_topic\} value aligns with the peak \{y\_axis\_topic\} of \{legend\_name\}?
\item What is the corresponding \{x\_axis\_topic\} value when \{legend\_name\} reaches its highest \{y\_axis\_topic\}?
\item When \{legend\_name\} reaches its highest \{y\_axis\_topic\}, what is the corresponding \{x\_axis\_topic\} value?
\item What \{x\_axis\_topic\} value corresponds to the highest \{y\_axis\_topic\} for \{legend\_name\}?
\item When \{legend\_name\} has its highest \{y\_axis\_topic\}, what is the corresponding \{x\_axis\_topic\} value?
\item What is the \{x\_axis\_topic\} value at the highest \{y\_axis\_topic\} for \{legend\_name\}?
\item When \{legend\_name\} hits its highest \{y\_axis\_topic\}, what is the corresponding \{x\_axis\_topic\} value?
\item When \{legend\_name\} reaches its highest point, what is the corresponding \{x\_axis\_topic\} value?
\item When \{legend\_name\} is at its highest point, what is the corresponding \{x\_axis\_topic\} value?
\item What is the \{x\_axis\_topic\} value when \{legend\_name\} reaches its highest point?
\item At the highest point of \{legend\_name\}, what is the corresponding \{x\_axis\_topic\} value?
\item What \{x\_axis\_topic\} value corresponds to the highest point of \{legend\_name\}?
\item When \{legend\_name\} reaches its peak, what is the corresponding \{x\_axis\_topic\} value?
\item What is the \{x\_axis\_topic\} value of the \{legend\_name\} with the lowest \{y\_axis\_topic\} in this chart?
\item In this chart, what \{x\_axis\_topic\} value corresponds to the \{legend\_name\} with the minimum \{y\_axis\_topic\}?
\item For the \{legend\_name\} with the lowest \{y\_axis\_topic\} in this chart, what is the \{x\_axis\_topic\} value?
\item What \{x\_axis\_topic\} value is associated with the \{legend\_name\} that has the lowest \{y\_axis\_topic\} in this chart?
\item Which \{x\_axis\_topic\} value corresponds to the lowest \{y\_axis\_topic\} for the \{legend\_name\} in this chart?
\item In this chart, what is the \{x\_axis\_topic\} value for the \{legend\_name\} with the smallest \{y\_axis\_topic\}?
\item What is the corresponding \{x\_axis\_topic\} value when \{legend\_name\} reaches its lowest \{y\_axis\_topic\}?
\item When \{legend\_name\} has its lowest \{y\_axis\_topic\}, what is the corresponding \{x\_axis\_topic\} value?
\item What \{x\_axis\_topic\} value corresponds to the lowest \{y\_axis\_topic\} for \{legend\_name\}?
\item For \{legend\_name\} at its minimum \{y\_axis\_topic\}, what is the corresponding \{x\_axis\_topic\} value?
\item Which \{x\_axis\_topic\} value aligns with \{legend\_name\} when it has the lowest \{y\_axis\_topic\}?
\item When the \{y\_axis\_topic\} is at its lowest for \{legend\_name\}, what is the corresponding \{x\_axis\_topic\} value?
\item When \{legend\_name\} reaches its lowest point, what is the corresponding \{x\_axis\_topic\} value?
\item What is the \{x\_axis\_topic\} value when \{legend\_name\} hits its lowest point?
\item At the lowest point of \{legend\_name\}, what is the corresponding \{x\_axis\_topic\} value?
\item What \{x\_axis\_topic\} value corresponds to the lowest point of \{legend\_name\}?
\item When \{legend\_name\} is at its lowest, what is the \{x\_axis\_topic\} value?
\item What \{x\_axis\_topic\} value aligns with the lowest point of \{legend\_name\}?
\item Which legend has a \{y\_axis\_topic\} equal to \{y\_value\} when the \{x\_axis\_topic\} is \{x\_value\}?
\item Which legend shows a \{y\_axis\_topic\} of \{y\_value\} when the \{x\_axis\_topic\} equals \{x\_value\}?
\item When the \{x\_axis\_topic\} is \{x\_value\}, which legend corresponds to a \{y\_axis\_topic\} of \{y\_value\}?
\item At \{x\_value\} on the \{x\_axis\_topic\}, which legend has a \{y\_axis\_topic\} value of \{y\_value\}?
\item What legend's \{y\_axis\_topic\} is \{y\_value\} when the \{x\_axis\_topic\} reads \{x\_value\}?
\item When \{x\_value\} is the value of the \{x\_axis\_topic\}, which legend has a \{y\_axis\_topic\} of \{y\_value\}?
\item For an \{x\_axis\_topic\} of \{x\_value\}, which legend displays a \{y\_axis\_topic\} of \{y\_value\}?
\item Which legend has a value of \{y\_value\} when the \{x\_axis\_topic\} is \{x\_value\}?
\item Which legend shows a value of \{y\_value\} when the \{x\_axis\_topic\} equals \{x\_value\}?
\item When the \{x\_axis\_topic\} is \{x\_value\}, which legend corresponds to a value of \{y\_value\}?
\item At \{x\_value\} on the \{x\_axis\_topic\}, which legend has a value of \{y\_value\}?
\item What legend's value is \{y\_value\} when the \{x\_axis\_topic\} reads \{x\_value\}?
\item When \{x\_value\} is the value of the \{x\_axis\_topic\}, which legend has a value of \{y\_value\}?
\item For an \{x\_axis\_topic\} of \{x\_value\}, which legend displays a value of \{y\_value\}?
\end{itemize}

\subsection*{EvoChart-QA Source Websites}
We selected 650 images from 140 publicly available websites for academic research purposes only. All sources are listed as follows:

\begin{itemize}
\small
\item https://www.beautiful.ai
\item https://www.formsbirds.com
\item https://leanscape.io
\item https://www.investopedia.com
\item https://www.storytellingwithdata.com
\item https://blog.finxter.com
\item https://www.degruyter.com
\item https://www.anychart.com
\item https://www.infragistics.com
\item https://awesomeopensource.com
\item https://fluttercore.com
\item https://www.nicesnippets.com
\item http://20bits.com
\item https://unreasonablegroup.com
\item https://mavink.com
\item https://www.smartsheet.com
\item https://template.wps.com
\item https://learn.microsoft.com
\item https://www.zoho.com
\item https://keski.condesan-ecoandes.org
\item https://www.statmethods.net
\item https://www.theinformationlab.com
\item https://www.pluralsight.com
\item https://www.visualitics.it
\item https://dribbble.com
\item https://infogram.com
\item https://beautifulai-od3.appspot.com 
\item https://www.slideteam.net
\item https://sainsdata.id 
\item https://www.elegantthemes.com
\item https://www.polymersearch.com
\item https://blog.csdn.net
\item https://aten.edu.vn
\item https://www.sakuranpost.net
\item https://imagesee.biz
\item https://search.justgulfwon.live
\item https://p.codekk.com
\item https://vitalflux.com
\item https://zebrabi.com
\item https://classfullprecisions.z13.web.core.windows.net
\item https://www.infocaptor.com
\item https://www.monkeybreadsoftware.de
\item https://mdpi.com
\item https://www.calxa.com
\item https://byggipedia.se
\item https://www.template.net
\item https://www.devtodev.com 
\item https://www.bakertilly.com
\item https://www.researchgate.net 
\item https://www.tessresearch.org
\item https://www.tandfonline.com
\item http://ww25.chartexamples.com
\item https://www.exceldemy.com
\item https://in.pinterest.com 
\item https://blog.51cto.com
\item https://www.fusioncharts.com
\item https://inforiver.com
\item https://exceljet.net
\item https://x.com
\item https://stevenrattner.com 
\item https://www.tillerhq.com
\item https://knifeknowitall.com
\item https://exploratory.io 
\item https://www.r-bloggers.com
\item https://jethrojeff.com/
\item https://marcuscalan.blogspot.com
\item https://www.pewresearch.org/
\item https://www.smashingmagazine.com
\item https://chart-studio.plotly.com
\item https://www.mdpi.com
\item https://loganix.com
\item https://www.knowbe4.com
\item https://www.zdnet.com
\item https://goldhartmediation.ca
\item https://wiseinvestments.ca
\item https://laptrinhx.com
\item https://mungfali.com
\item https://data-flair.training
\item https://www.ncl.ac.uk
\item https://www.dummies.com
\item https://georgecarlo.blogspot.com
\item http://www.aploris.com
\item https://respect.international
\item https://www.pinterest.com
\item https://www.educba.com
\item https://www.statista.com/
\item https://slidebazaar.com
\item https://venngage.com
\item https://airfreesm.best
\item https://ar.pinterest.com
\item https://www.conceptdraw.com
\item https://www.ft.com
\item https://www.statology.org
\item https://exceljet.net/charts
\item https://www.slidekit.com
\item https://worksheetsploshes.z14.web.core.windows.net
\item https://www.hotzxgirl.com
\item https://www.genekitr.fun
\item https://medium.com
\item https://daydreamingnumbers.com
\item https://www.newsweek.com
\item https://www.mekkographics.com
\item https://nowbam.com
\item https://www.mercurynews.com
\item https://www.everviz.com 
\item https://byjus.com
\item https://www.dailyrecord.co.uk
\item https://appfire.com
\item https://www.aiophotoz.com
\item https://socialbarrel.com
\item https://www.canadianpizzamag.com
\item https://edubenchmark.com
\item https://ieltsessentialindia.blogspot.com
\item https://forum.knime.com 
\item https://docs.oracle.com
\item https://www.cec.health.nsw.gov.au
\item https://www.linkedin.com 
\item https://online.stat.psu.edu
\item https://gmt-tutorials.org
\item https://gitcode.csdn.net
\item https://online.visual-paradigm.com 
\item https://python-charts.com
\item https://cloud.tencent.com
\item https://www.cnblogs.com
\item https://help.xlstat.com
\item https://www.listendata.com
\item https://docs.thoughtspot.com
\item https://www.data-to-viz.com
\item https://fahimahmad.netlify.app 
\item https://developer.aliyun.com
\item https://visme.co
\item https://pythonspot.com
\item https://data36.com
\item https://bootcamp.uxdesign.cc
\item https://revistaplural.es
\item https://environicsanalytics.com
\item https://mainpackage9.gitlab.io
\item https://en.wikipedia.org
\item https://riset.guru.pubiway.com
\item https://lopezcollege.weebly.com
\end{itemize}

\end{document}